\documentclass[10pt,twocolumn,letterpaper]{article}

\usepackage{cvpr}      % To produce the REVIEW version
\usepackage{graphicx}
\usepackage{amsmath}
\usepackage{amssymb}
\usepackage{booktabs, array}
\usepackage{multirow}
\usepackage{comment}
\usepackage{color, colortbl}
\usepackage{times}
\usepackage{epsfig}
\usepackage{graphicx}
\usepackage{amsmath}
\usepackage{amssymb}
\usepackage[percent]{overpic}
\usepackage{times}
\usepackage{epsfig}
\usepackage{graphicx}
\usepackage{float}
\usepackage{wrapfig}
\usepackage{amsmath,amssymb,amsthm}
\usepackage{algorithm,algorithmicx,algpseudocode}
\usepackage{bm,xspace}
\usepackage{comment}
\usepackage{verbatim}
\usepackage{multirow}
\usepackage{balance}
\usepackage{url}
\usepackage{booktabs}
\usepackage{etoolbox,siunitx}
\usepackage{calc}
\usepackage{pifont,hologo}
\usepackage{dblfloatfix}
\usepackage{soul}
\usepackage{bbding}
\usepackage{pifont}
\usepackage{wasysym}

\usepackage{bbm}

\usepackage{arydshln}
\usepackage[accsupp]{axessibility}

\definecolor{gold}{rgb}{1.0, 0.874, 0}
\definecolor{silver}{rgb}{0.77,0.77,0.77}
\definecolor{brown}{rgb}{0.95, 0.678, 0.4}

\newcommand{\gold}[1]{\colorbox{gold}{\textbf{#1}}}
\newcommand{\silver}[1]{\colorbox{silver}{\textbf{#1}}}
\newcommand{\bronze}[1]{\colorbox{brown}{\textbf{#1}}}

\usepackage[breaklinks,colorlinks]{hyperref}

\usepackage[capitalize]{cleveref}
\crefname{section}{Sec.}{Secs.}
\Crefname{section}{Section}{Sections}
\Crefname{table}{Table}{Tables}
\crefname{table}{Tab.}{Tabs.}

\definecolor{LightYellow}{rgb}{1,0.7,0.7}
\definecolor{LightPink}{rgb}{0.7,1,0.7}
\definecolor{cream}{rgb}{1.0, 0.99, 0.82}
\def\cvprPaperID{6804} % *** Enter the CVPR Paper ID here
\def\confName{CVPR}
\def\confYear{2023}
\begin{document}

%%%%%%%%% TITLE - PLEASE UPDATE
%\title{Self-Supervised Deep Stereo and Optical Flow leveraging 3D Nerf Supervision}

\title{NeRF-Supervised Deep Stereo}

\author{Fabio Tosi$^1$ \quad\quad Alessio Tonioni$^2$ \quad\quad Daniele De Gregorio$^3$ \quad\quad Matteo Poggi$^1$\\
$^1$University of Bologna \quad\quad $^2$Google Inc. \quad\quad $^3$Eyecan.ai\\
{\tt\small \{fabio.tosi5, m.poggi\}@unibo.it} \hspace{0.3cm} {\tt\small alessiot@google.com} \hspace{0.3cm} {\tt\small daniele.degregorio@eyecan.ai}\\
\small\url{https://nerfstereo.github.io/}
% For a paper whose authors are all at the same institution,
% omit the following lines up until the closing ``}''.
% Additional authors and addresses can be added with ``\and'',
% just like the second author.
% To save space, use either the email address or home page, not both
}

%\maketitle

%%%%%%%%% BODY TEXT

\twocolumn[{
\renewcommand\twocolumn[1][]{#1}
\maketitle
\begin{center}
    \vspace{-0.8cm}
    \setlength{\tabcolsep}{1.5pt}
    \includegraphics[trim=0cm 15cm 16cm 0cm ,clip,width=\textwidth]{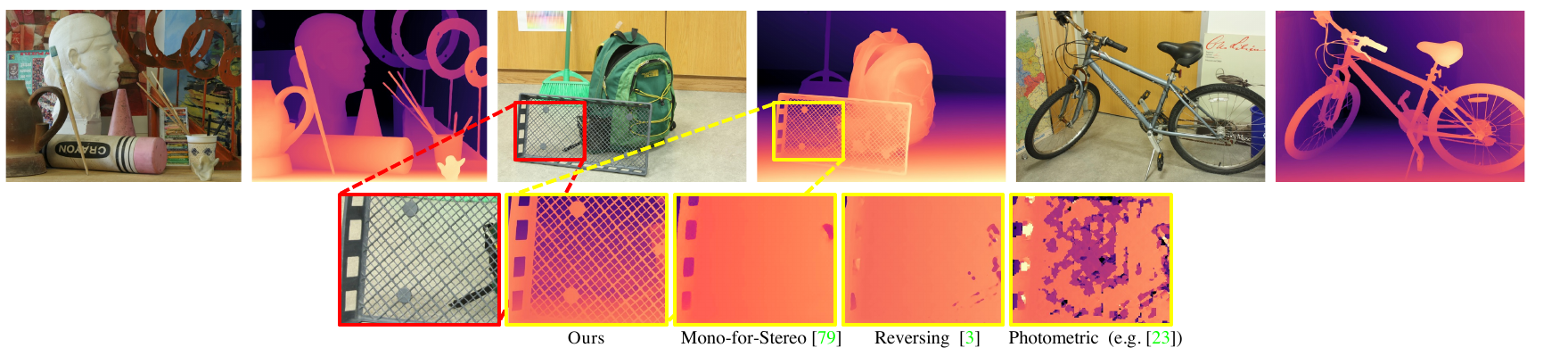}
    \label{fig:teaser}
\end{center}
\vspace{-0.8cm}
\small \hypertarget{fig:teaser}{Fig. 1.} \textbf{Zero-Shot Generalization Results.} On top, predictions by RAFT-Stereo \cite{lipson2021raft} trained with our approach on user-collected images, without using any synthetic datasets, ground-truth depth or (even) real stereo pairs. At the bottom, a zoom-in over the \textit{Backpack} disparity map, showing an unprecedented level of detail compared to existing strategies not using ground-truth \cite{watson2020stereo,aleotti2020reversing,Godard_CVPR_2017} trained with the same data.
\vspace{0.2cm}
}]
    
\begin{abstract}
We introduce a novel framework for training deep stereo networks effortlessly and without any ground-truth. By leveraging state-of-the-art neural rendering solutions, we generate stereo training data from image sequences collected with a single handheld camera. On top of them,  a NeRF-supervised training procedure is carried out, from which we exploit rendered stereo triplets to compensate for occlusions and depth maps as proxy labels. This results in stereo networks capable of predicting sharp and detailed disparity maps. Experimental results show that models trained under this regime yield a 30-40\% improvement over existing self-supervised methods on the challenging Middlebury dataset, filling the gap to supervised models and, most times, outperforming them at zero-shot generalization.
\end{abstract}
\vspace{-0.5cm}

%%%%%%%%% BODY TEXT

%-------------------------------------------------------------------------

\section{Introduction}

Depth from stereo is one of the longest-standing research fields in computer vision \cite{marr1976cooperative}. It involves finding pixels correspondences across two \textit{rectified} images to obtain the disparity -- i.e., their difference in terms of horizontal coordinates -- and then use it to triangulate depth.
After years of studies with hand-crafted algorithms \cite{scharstein2002taxonomy}, deep learning radically changed the way of approaching the problem \cite{zbontar2016stereo}.
End-to-end deep networks \cite{poggi2021synergies} rapidly became the dominant solution for stereo, delivering outstanding results on benchmarks \cite{Menze2015CVPR,scharstein2014high,schops2017multi} given sufficient training data.

This latter requirement is the key factor for their success, but it is also one of the greatest limitations. Annotated data is hard to source when dealing with depth estimation since additional sensors are required (e.g., LiDARs), and thus represents a thick entry barrier to the field. Over the years, two main trends have allowed to soften this problem: self-supervised learning paradigms \cite{Godard_CVPR_2017,wang2019unos,aleotti2020reversing} and the use of synthetic data \cite{mayer2016large,tartanair2020iros,li2022practical}.
Despite these advances, both approaches still have weaknesses to address.

\textit{Self-supervised learning:} despite the possibility to train on any unlabeled stereo pair collected by any user -- and potentially opening to data {democratization} --
the use of self-supervised losses is ineffective at dealing with ill-posed stereo settings (e.g. occlusions, non-Lambertian surfaces, etc.). Albeit recent approaches soften the occlusions problem \cite{aleotti2020reversing}, predictions are far from being as sharp, detailed and accurate as those obtained through supervised training. Moreover, the self-supervised stereo literature \cite{wang2019unos,lai2019bridging,chen2021revealing} often focuses on well-defined domains (i.e., KITTI) and rarely exposes domain generalization capabilities \cite{aleotti2020reversing}.

\textit{Synthetic data:} although training on densely annotated synthetic images can guide the networks towards sharp, detailed and accurate predictions, the {domain-shift} that occurs when testing on real data dampens the full potential of the trained model. A large body of recent literature addressing zero-shot generalization \cite{aleotti2021neural,lipson2021raft,li2022practical,liu2022graftnet,chuah2022itsa,zhang2022revisiting} proves how relevant the problem is. 
However, obtaining stereo pairs as realistic as possible requires significant effort, despite synthetic depth labels being easily sourced through a graphics rendering pipeline. Indeed, modelling high-quality assets is crucial for mitigating the domain shift and requires excellent graphics skills. While artists make various assets available, these are seldom open source and necessitate additional human labor to be organized into plausible scenes.

%\rule{0pt}{3ex}  
In short, in a world where data is the new gold, obtaining flexible and scalable training samples to unleash the full potential of deep stereo networks still remains an open problem. In this paper, we propose a novel paradigm to address this challenge.
Given the recent advances in neural rendering \cite{nerf,mueller2022instant}, we exploit them as \textit{data factories}: we collect sparse sets of images in-the-wild with a standard, single handheld camera. After that, we train a \textit{Neural Radiance Field} (NeRF) model for each sequence and use it to render arbitrary, novel views of the same scene. Specifically, we synthesize {stereo pairs} from arbitrary viewpoints by rendering a \textit{reference} view corresponding to the real acquired image, and a \textit{target} one on the right of it, displaced by means of a virtual arbitrary baseline. This allows us to generate countless samples to train any stereo network in a self-supervised manner by leveraging popular photometric losses \cite{Godard_CVPR_2017}. However, this na\"ive approach would inherit the limitations of self-supervised methods \cite{wang2019unos,lai2019bridging,chen2021revealing} at occlusions, which can be effectively addressed by rendering a \textit{third} view for each pair, placed on the left of the source view specularly to the other target image. This allows to compensate for the missing supervision at occluded regions. Moreover, proxy-supervision in the form of rendered depth by NeRF completes our \textit{NeRF-Supervised} training regime. 
With it, we can train deep stereo networks by conducting a low-effort collection campaign, and yet obtain state-of-the-art results without requiring any ground-truth label -- or not even a real stereo camera! -- as shown on top of Fig. \hyperref[fig:teaser]{1}.

We believe that our approach is a significant step towards democratizing training data. In fact, we will demonstrate how the efforts of just the four authors were enough to collect sufficient data (roughly 270 scenes) to allow our NeRF-Supervised stereo networks to outperform models trained on synthetic datasets, such as \cite{aleotti2021neural,lipson2021raft,li2022practical,liu2022graftnet,chuah2022itsa,zhang2022revisiting}, as well as existing self-supervised methods \cite{watson2020stereo,aleotti2020reversing} in terms of zero-shot generalization, as depicted at the bottom of Fig. \hyperref[fig:teaser]{1}.

We summarize our main contributions as:

\begin{itemize}
    \item A novel paradigm for collecting and generating stereo training data using neural rendering and a collection of user-collected image sequences.
    \item A NeRF-Supervised training protocol that combines rendered image triplets and depth maps to address occlusions and enhance fine details. 
    \item State-of-the art, zero-shot generalization results on challenging stereo datasets \cite{scharstein2014high}, without exploiting any ground-truth or real stereo pair.
\end{itemize}
%To conclude, in case of acceptance, we will release our self-collected dataset, as well as launch a web platform allowing any users to upload their own image collections and to help into training better and better stereo networks, as well as to put their own brick on the road for full data democratization.

%-------------------------------------------------------------------------

\begin{figure*}[t]
    \centering
    \includegraphics[ width=0.95\linewidth]{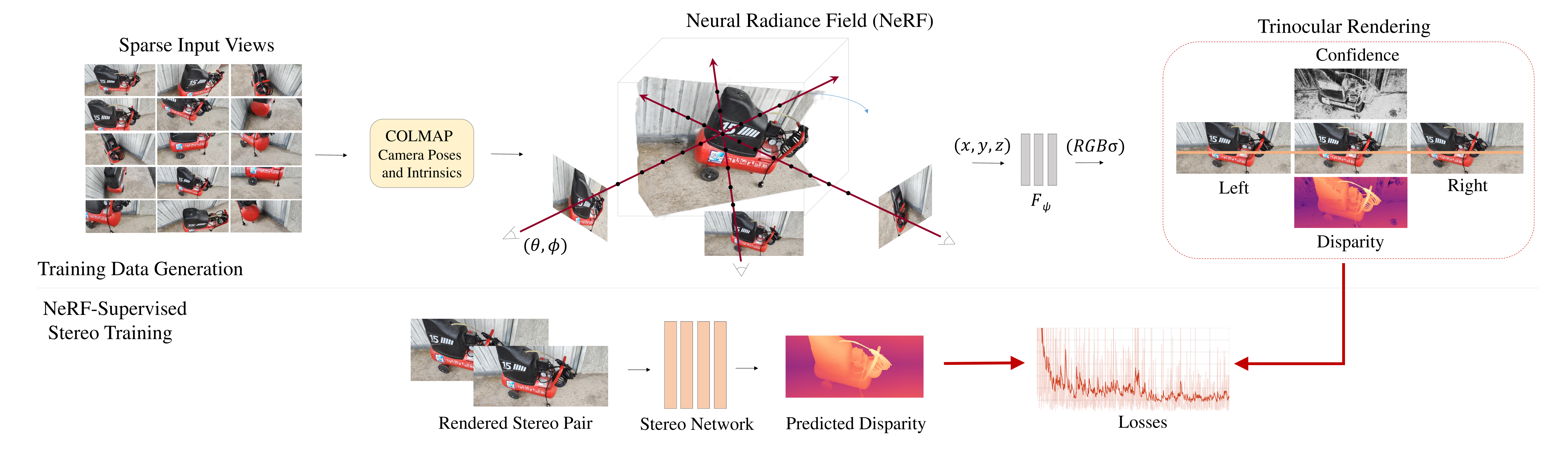}
    \vspace{-0.45cm}\caption{\textbf{ Framework Overview.} Top: data generation pipeline that trains NeRFs from user-collected single-camera frames to render stereo images (i.e. triplets), confidence and proxy depth maps. Bottom:  NeRF-supervised training of a stereo network on rendered pairs.
    }\vspace{-0.3cm}
    \label{fig:framework}
\end{figure*}

\section{Related work}

\textbf{Deep Stereo Matching.} For decades, stereo matching has been tackled using hand-crafted algorithms \cite{scharstein2002taxonomy} usually classified into local and global methods, according to their processing step and their speed/accuracy trade-off.
In recent years, deep learning has become the dominant technique in the stereo matching field, achieving results that were previously unthinkable  \cite{poggi2021synergies}. Early efforts in this field cast individual steps of the pipeline \cite{scharstein2002taxonomy} as learnable components \cite{zbontar2016stereo,luo2016efficient,seki2016patch,seki2017sgm-net}. Starting with DispNet \cite{mayer2016large}, end-to-end architectures rapidly replaced any alternative approach \cite{Kendall_2017_ICCV,Liang_2018_CVPR,chang2018psmnet,zhang2019ga,yang2019hierarchical,cheng2020hierarchical,zhang2019domaininvariant,Tosi2021CVPR,shen2021cfnet}.
The latest advances in this field take inspiration from RAFT \cite{teed2020raft} to design recurrent architectures, either by performing lookups on a 3D correlation volume \cite{lipson2021raft} or correlations in a local window \cite{li2022practical}, or exploiting Transformers \cite{li2021revisiting,guo2022context} to capture long-range dependencies between features of the input stereo pair.
Despite their impressive results on public benchmarks, these methods strictly require dense in-domain ground-truth.

\textbf{Self-Supervised Stereo.} This branch of stereo literature aims to train deep models without the use of ground-truth depth data.
A common strategy involves using photometric losses \cite{Godard_2019_ICCV} across stereo images from single pairs \cite{SsSMnet2017,Tonioni_2019_CVPR,Tonioni_2019_learn2adapt} or videos \cite{lai2019bridging,wang2019unos,chi2021feature}. 
An alternative line of works replaces it with proxy supervision from either hand-crafted algorithms \cite{Tonioni_2017_ICCV,tonioni2020unsupervised,Poggi2021continual} or distilled from other networks \cite{aleotti2020reversing}. Although these strategies are practical, they have proven to only be effective at specializing or adapting to single domains, and often lack generalization \cite{aleotti2020reversing}, yet not providing reliable supervision at occlusions. In contrast, we exploit multi-view geometry at its finest through neural rendering to learn for stereo, alike single-image depth estimation frameworks can learn from stereo images \cite{Godard_CVPR_2017}.

\textbf{Zero-Shot Generalization.} 
This line of work focuses on training deep models on a set of labeled images and then preserving accuracy when tested across different domains, under the assumption that target domain-specific data is unavailable.
Approaches initially explored include: the use of learning domain-invariant features \cite{zhang2019domaininvariant}, hand-crafted matching volumes \cite{cai2020matchingspace}, or casting disparity estimation as a refinement problem on top of hand-crafted stereo algorithms \cite{aleotti2021neural}. 
The latest trends in the field include using contrastive feature loss and stereo selective whitening loss \cite{zhang2022revisiting}, an ImageNet pre-trained classifier to extract general-purpose image features and graft them into the cost volume \cite{liu2022graftnet}, or shortcut avoidance \cite{chuah2022itsa}.
Among others, Mono-for-Stereo (MfS) \cite{watson2020stereo} generates training stereo pairs from large-scale real-world monocular datasets. This at the expense of 1) requiring a pre-trained monocular depth network \cite{Ranftl2022} -- which in turns is typically trained on \textit{millions} of images involving also ground-truth labels -- and 2) deal with holes generated by the forward-warping operation used to obtain the right view. 
In contrast, our approach of generating stereo pairs from single images does not require any model pre-trained on million images \cite{Ranftl2022}, ground-truth label or post-processing step, and still achieves better results. 

\textbf{Neural Radiance Fields.} NeRFs \cite{nerf} belong to the family of neural fields \cite{neuralfields}. These models implicitly parameterize a 5D lightfield using one or more Multi-Layer Perceptrons (MLPs). In just three years, they have become the dominant approach for generating novel views using neural rendering. Different flavors of NeRFs have been developed to deal with
dynamic scenes~\cite{Martin-BruallaR21,PumarolaCPM21,LiNSW21,XianHK021,GaoSKH21}, image relighting~\cite{SrinivasanDZTMB21,ZhangSDDFB21,BossBJBLL21}, camera poses refinement \cite{lin2021barf,wang2021nerf}, anti-aliasing in multi-resolution images \cite{Barron_2021_ICCV,barron2022mip}, cross-spectral imaging \cite{xnerf}, deformable objects~\cite{ParkSBBGSM21,TretschkTGZLT21,GafniTZN21,NoguchiSLH21,ParkSHBBGMS21} or content generation~\cite{SchwarzLN020,ChanMK0W21,KosiorekSZMSMR21}. Most recent NeRF variants focus on faster convergence, e.g. by exploiting multiple MLPs\cite{reiser2021kilonerf}, factorization \cite{Chen2022ECCV} or explicit representations~\cite{yu2021plenoxels,sun2021direct,mueller2022instant}.

Recent works partially explored the potential of NeRFs to serve as data factories at high-level -- object detection \cite{ge2022neural}, semantic labeling \cite{semanticnerf} or to learn descriptors \cite{yen2022nerfsupervision}.

%-------------------------------------------------------------------------

\section{Method}

 Fig. \ref{fig:framework} illustrates our NeRF-Supervised (NS) learning framework. We first collect multi-view images from multiple static scenes. 
 Then, we fit a NeRF on each single scene to render stereo \textit{triplets} and depth. 
Finally, the rendered data is used to train any existing stereo matching network.

\subsection{Background: Neural Radiance Field (NeRF)}
\label{sec:nerf}

A Neural Radiance Field (NeRF) \cite{nerf} maps a 5D input -- 3D coordinates $\mathbf{x} = (x,y,z)$ of a point in the scene and viewing directions  $(\theta,\phi)$ of the camera capturing it -- into a color-density output $(\mathbf{c}, \sigma)$ by means of a network $F_{\psi}$, modelling the radiance of an observed scene as $F_\psi(\mathbf{x}, \theta, \phi) \rightarrow (\mathbf{c}, \sigma)$.
Such a 5D function is approximated by the weights of an MLP $F_{\psi}$. 
To render a 2D image, the following steps are taken: 1) sending camera rays through the scene to sample a set of points, 2) estimate density and color for each sampled point with $F_{\psi}$ and 3) exploit volume rendering \cite{Max1995Optical} to synthesize the 2D image. In practice, the color $C(\mathbf{r})$ rendered from a camera ray $\mathbf{r}(t) = \mathbf{o} + t\mathbf{d}$ can be obtained by solving the following integral:

\begin{equation}\label{eq:rendering}
    C(\mathbf{r}) = \int_{t_n}^{t_f} T(t) \sigma(\mathbf{r}(t)) c(\mathbf{r}(t),\mathbf{d}) \textit{dt} 
\end{equation}
with $T(t) = exp \left( -\int_{t_n}^{t} \sigma (\textbf{r}(s))ds\right)$ representing the accumulated transmittance from $t_n$ to $t_f$ along the ray $r$, and $t_n, t_f$ the near and far plane, respectively.  
The integral is computed via quadrature by dividing the ray into a pre-defined set of N evenly spaced bins: 

\begin{equation}\label{eq:quadrature}
    C(\mathbf{r}) = \sum_{i=1}^{N} T_i(1-\text{exp}(-\sigma_i\delta_i))c_i, \hspace{0.4cm} T_i = \text{exp}\Big(-{\sum_{j=1}^{i-1}\sigma_j\delta_j}\Big)   
\end{equation}
with $\delta_i$ being the step between adjacent samples $t_i, t_{i+1}$.

\textbf{Speed-up with Explicit Representations.} The described model is effective, but slow to train due to two reasons: first, the MLP must learn from scratch the mapping for all points in the 5D space and second, for each individual input, the entire set of weights needs to be optimized.
Explicit representations -- e.g. voxel grids -- can store additional features that can be rapidly indexed and interpolated, but this comes at the cost of higher memory requirements. This allows for 1) a shallower MLP, faster to converge and 2) a reduced number of parameters to optimize for each single input -- i.e., features on a voxel grid and the few parameters of the shallow MLP. For instance, on this principle, DVGO \cite{sun2021direct} builds two voxel grids $\mathbf{M}^{\text{(dens)}}$ and $\mathbf{M}^{\text{(feat)}}$, with the former modeling density and the latter storing features that are queried by the MLP $F_\psi$ to compute color:

\begin{align}
\sigma(\mathbf{x}) &= \text{interp}(\mathbf{x}, \mathbf{M}^{\text{(dens)}}) \\ 
\mathbf{c}(\mathbf{x},\mathbf{d}) &= F_{\psi}(\text{interp}(\mathbf{x}, \mathbf{M}^{\text{(feat)}}), \mathbf{x}, \mathbf{d})    
\end{align}
while Instant-NGP \cite{mueller2022instant} builds multi-resolution voxel grids accessed by means of index hashing:

\begin{equation}
    h(\mathbf{x}) = \Big( \bigoplus_{i=0}^d x_i\pi_i \Big) \mod{T}
\end{equation}
with $\bigoplus$ being bitwise XOR, $x_i$ -- with $i \in [0, d]$ -- single bits of the location index $\mathbf{x}$, $\pi_i$ unique, large prime numbers and $T$ the maximum amount of elements in the grid.

\begin{figure*}
    \centering
    \renewcommand{\tabcolsep}{1pt}
    \begin{tabular}{cccccc}

         \begin{overpic}[width=0.18\linewidth]{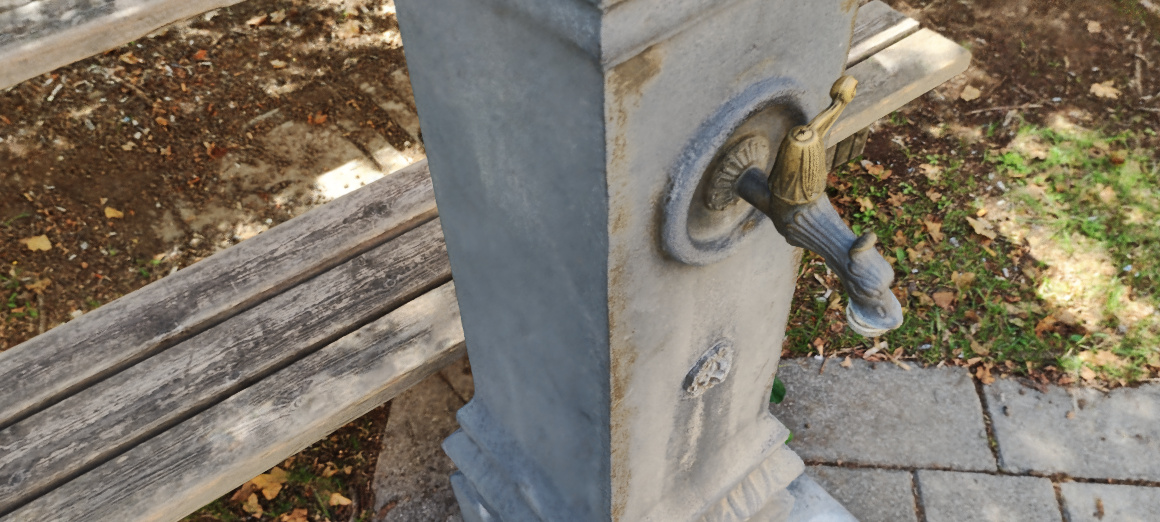}
         \put(0,38){\color{green}\line(2,0){104}}
         \put(0,16){\color{green}\line(2,0){104}}
         \put(0,4){\color{green}\line(2,0){104}}
         \put(0,38){\color{green}\line(2,0){104}}
         \put(0,16){\color{green}\line(2,0){104}}
         \put(0,4){\color{green}\line(2,0){104}}
         \put(0,38){\color{green}\line(2,0){104}}
         \put(0,16){\color{green}\line(2,0){104}}
         \put(0,4){\color{green}\line(2,0){104}}
         \put(0,2){\colorbox{white}{$\displaystyle\textcolor{black}{\text{(a)}}$}}
         \end{overpic} &
         \begin{overpic}[width=0.18\linewidth]{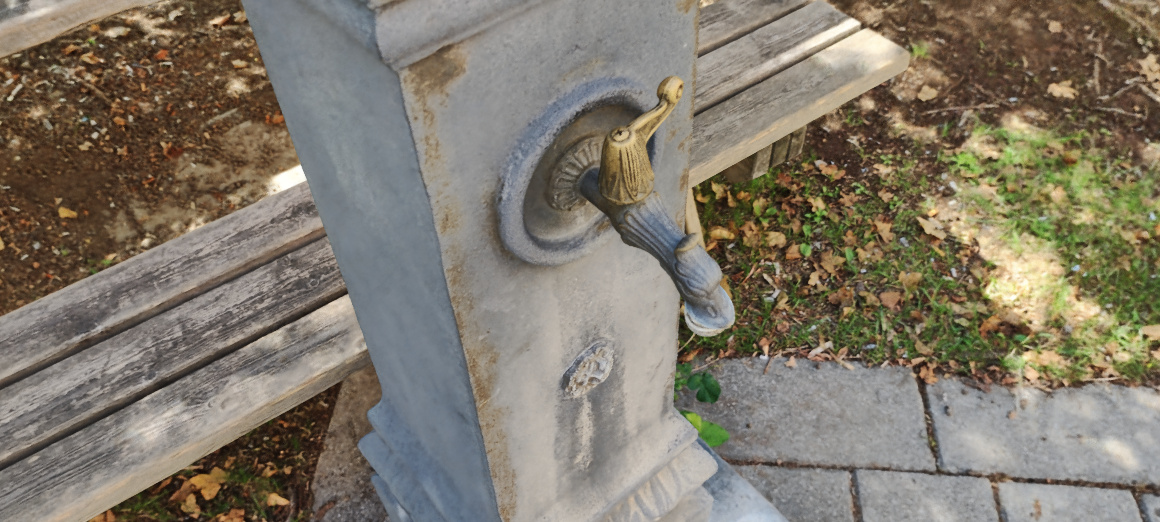}
         \put(0,38){\color{green}\line(2,0){104}}
         \put(0,16){\color{green}\line(2,0){104}}
         \put(0,4){\color{green}\line(2,0){104}}
         \put(0,38){\color{green}\line(2,0){104}}
         \put(0,16){\color{green}\line(2,0){104}}
         \put(0,4){\color{green}\line(2,0){104}}
         \put(0,38){\color{green}\line(2,0){104}}
         \put(0,16){\color{green}\line(2,0){104}}
         \put(0,4){\color{green}\line(2,0){104}}
         \put(0,2){\colorbox{white}{$\displaystyle\textcolor{black}{\text{(b)}}$}}
         \end{overpic} &
          \begin{overpic}[width=0.18\linewidth]{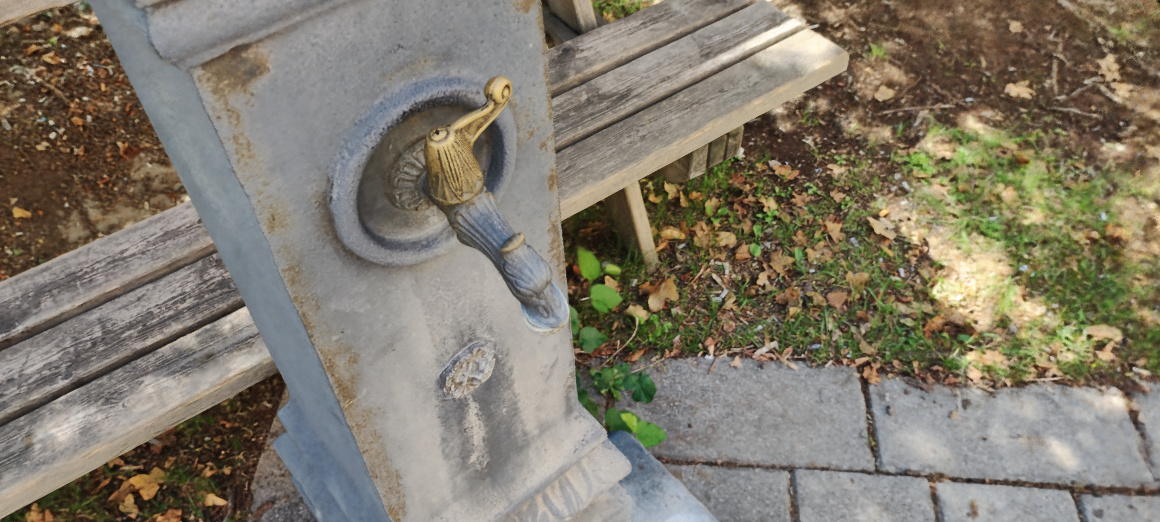}
         \put(0,38){\color{green}\line(2,0){100}}
         \put(0,16){\color{green}\line(2,0){100}}
         \put(0,4){\color{green}\line(2,0){100}}
         \put(0,38){\color{green}\line(2,0){100}}
         \put(0,16){\color{green}\line(2,0){100}}
         \put(0,4){\color{green}\line(2,0){100}}
         \put(0,38){\color{green}\line(2,0){100}}
         \put(0,16){\color{green}\line(2,0){100}}
         \put(0,4){\color{green}\line(2,0){100}}
         \put(0,2){\colorbox{white}{$\displaystyle\textcolor{black}{\text{(c)}}$}}
         \end{overpic} & \quad\quad\quad\quad &
         \begin{overpic}[width=0.18\linewidth]{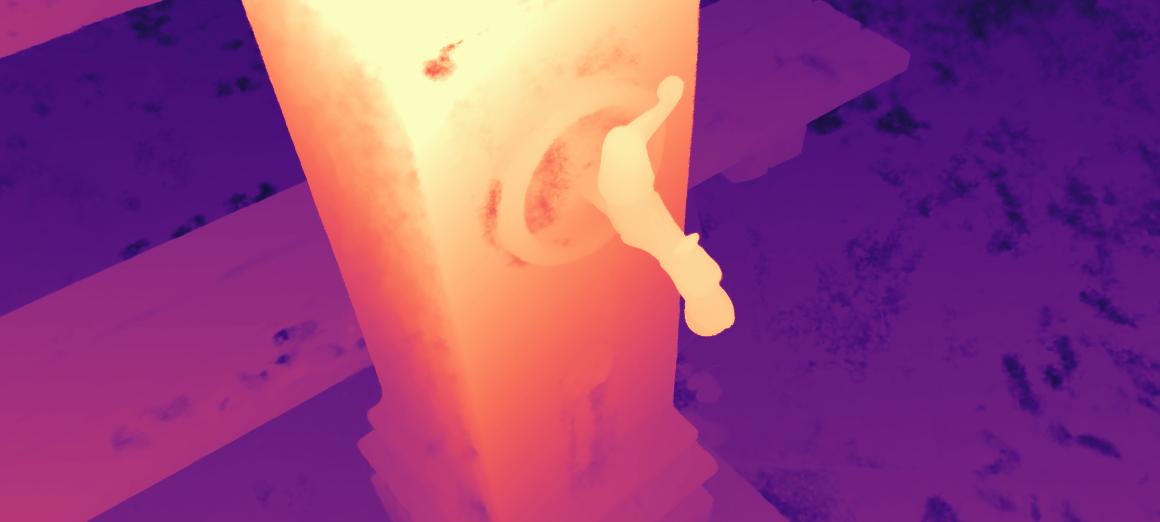}
         \put(0,2){\colorbox{white}{$\displaystyle\textcolor{black}{\text{(g)}}$}}
         \end{overpic} 
         &
         \begin{overpic}[width=0.18\linewidth]{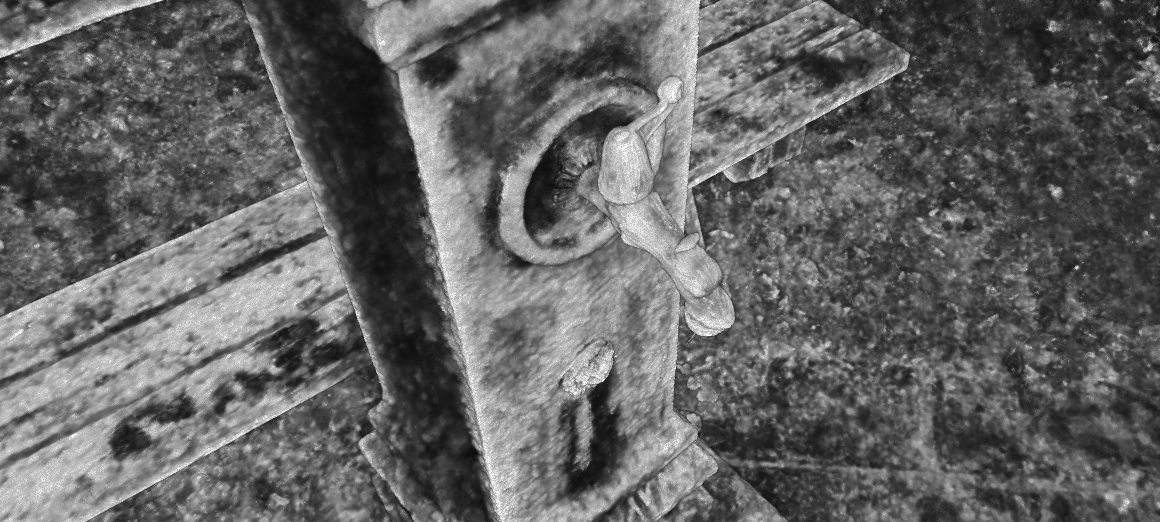}
         \put(0,2){\colorbox{white}{$\displaystyle\textcolor{black}{\text{(h)}}$}}
         \end{overpic} \\
         \begin{overpic}[width=0.18\linewidth]{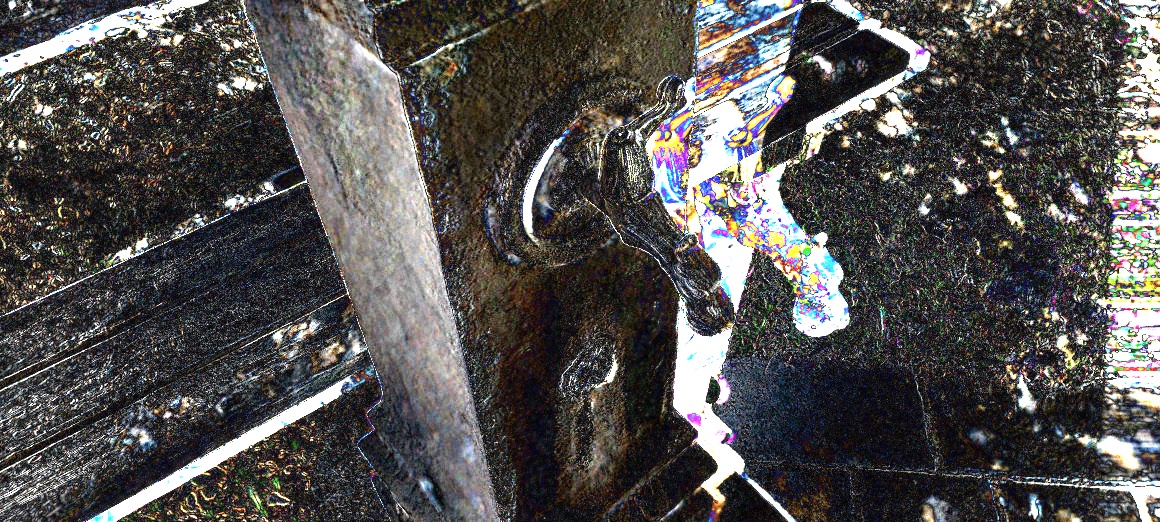}
         \put(0,2){\colorbox{white}{$\displaystyle\textcolor{black}{\text{(d)}}$}}
         \end{overpic} &
         \begin{overpic}[width=0.18\linewidth]{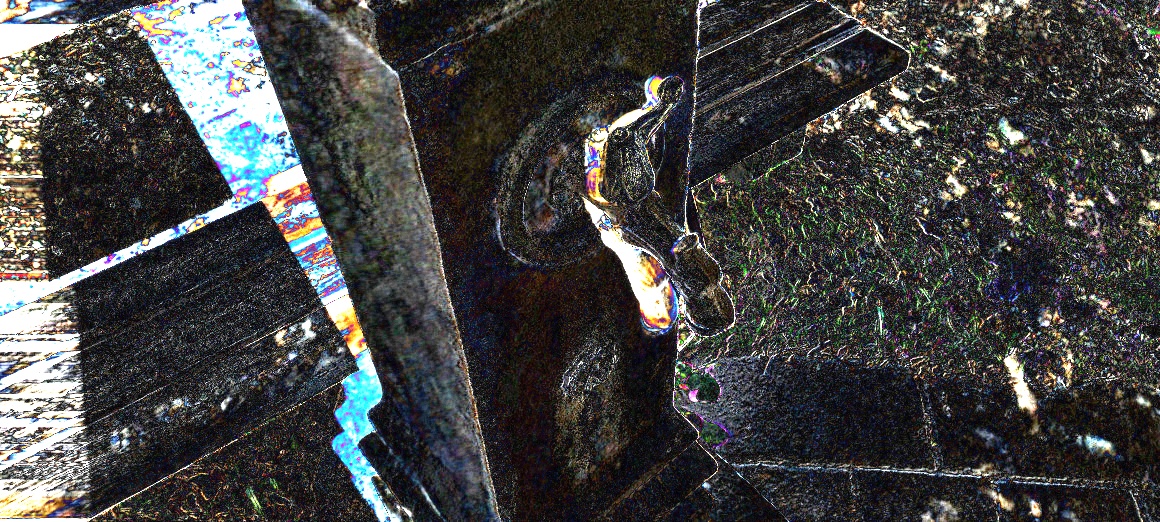}
         \put(0,2){\colorbox{white}{$\displaystyle\textcolor{black}{\text{(e)}}$}}
         \end{overpic} &
         \begin{overpic}[width=0.18\linewidth]{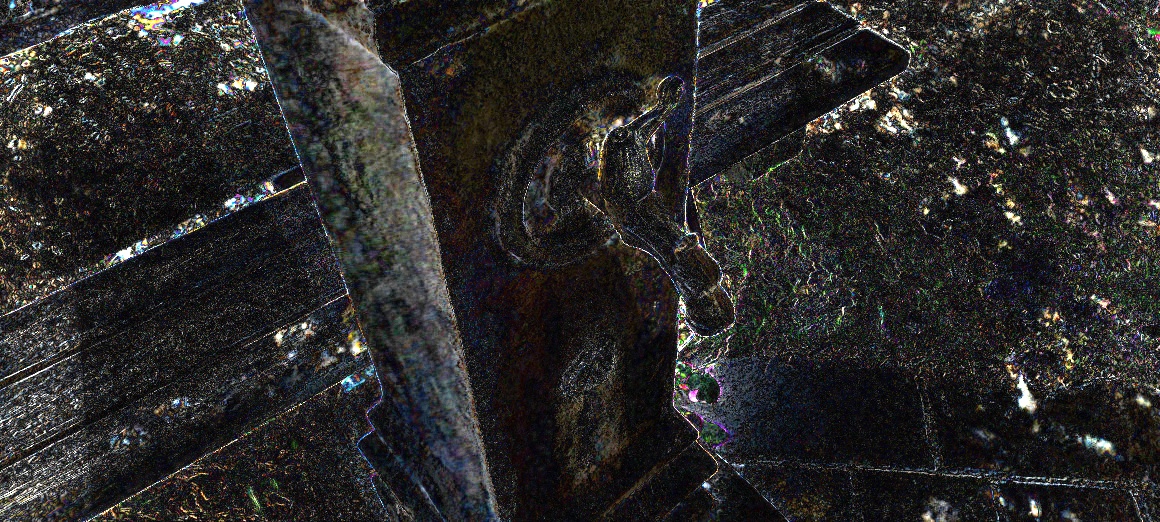}
         \put(0,2){\colorbox{white}{$\displaystyle\textcolor{black}{\text{(f)}}$}}
         \end{overpic} & \quad\quad\quad\quad &
         \begin{overpic}[width=0.18\linewidth]{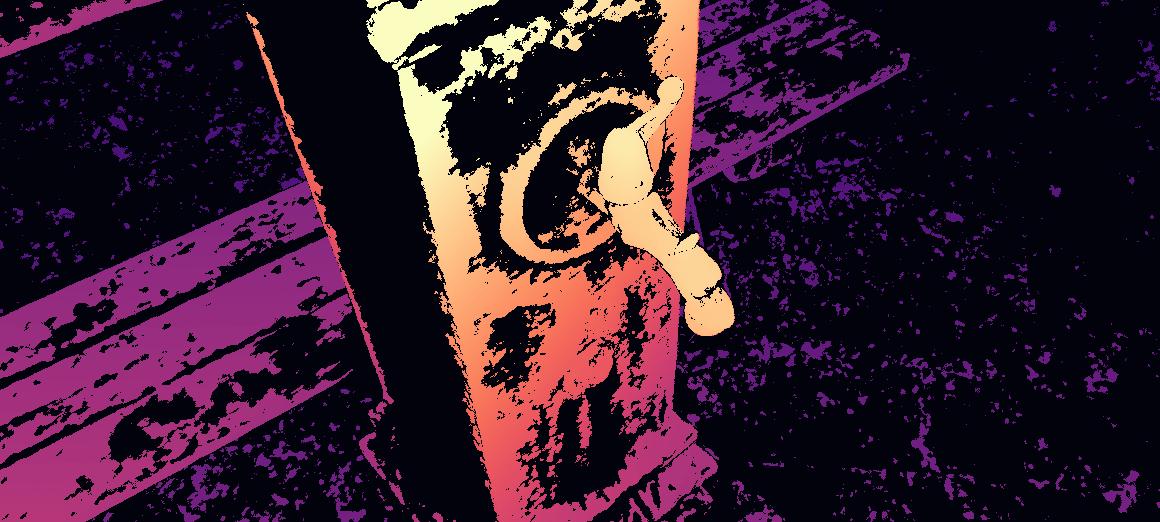}
         \put(0,2){\colorbox{white}{$\displaystyle\textcolor{black}{\text{(i)}}$}}
         \end{overpic} & \begin{overpic}[width=0.18\linewidth]{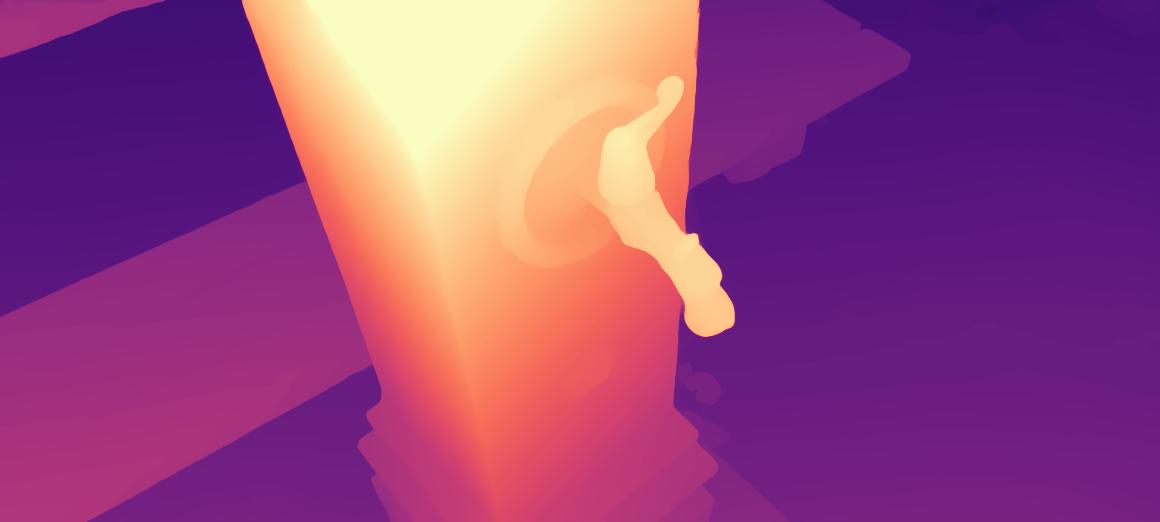}
         \put(0,2){\colorbox{white}{$\displaystyle\textcolor{black}{\text{(j)}}$}}\end{overpic} \\
    \end{tabular}
    \vspace{-0.3cm}
    \caption{\textbf{Visualization of NS loss components}. On the left: (a-c) rendered left-center-right triplet. (d) left-to-center and (e) right-to-center photometric losses, both exposing their own occlusions, (f) per-pixel minimum, compensating for them. On the right: (g) NeRF rendered noisy disparity map, (h) Ambient Occlusion (AO), (i) AO-filtered NeRF disparity, (j) prediction by RAFT-Stereo trained on our dataset. }\vspace{-0.3cm}
    \label{fig:losses}
\end{figure*}

\subsection{NeRF as a Data Factory}

With reference to Fig. \ref{fig:framework}, we now describe, step-by-step, how we use Neural Radiance Fields to generate countless image pairs for training any deep stereo network.

\textbf{Image Collection and COLMAP Pre-processing.} We start by acquiring a sparse set of M images from a single static scene, for which any handheld device with a single camera is suitable, e.g., a mobile phone. 
We run COLMAP \cite{schoenberger2016sfm} on any single scene to estimate both intrinsics $\mathbf{K}$ and camera poses $\mathbf{E}_i, i \in [1,\text{M}]$. This is a standard procedure for preparing user-collected data to be NeRFed \cite{nerf,Barron_2021_ICCV,barron2022mip}.

\textbf{NeRF Training.} Then, we fit an independent NeRF -- in one of its speeded-up flavors \cite{yu2021plenoxels,sun2021direct,mueller2022instant,Chen2022ECCV} -- for each scene. This is achieved by rendering, for a given batch $\mathcal{R}$ of rays shot from collected image positions, the corresponding color $\hat{C}(\mathbf{r})$ according to Eq. \ref{eq:quadrature}, and optimizing an L2 loss with respect to pixel colors $C(\mathbf{r})$ in the collected frames:

\begin{equation}
    \mathcal{L}_{rend} = \sum_{\mathbf{r} \in \mathcal{R}} ||\hat{C}(\mathbf{r}) - C(\mathbf{r}) ||_2^2
\end{equation}
%plus additional, implementation-specific terms \cite{rudin1994total,sun2021direct}.

\textbf{Stereo Pairs Rendering.} Finally, to create the stereo training set, we generate a virtual set of stereo extrinsic parameters $\mathbf{S} = \mathbb{I}|\mathbf{b}$. The rotation is represented by the $3\times3$ identity matrix $\mathbb{I}$ and the translation vector $\mathbf{b} = (b,0,0)^\text{T}$ has a magnitude $b$ along the $x$ axis in the camera reference system. This defines the baseline of a virtual stereo camera.

Subsequently, we render two novel views, one originating from an arbitrary viewpoint $\mathbf{E}_k = \mathbf{R}_k|\mathbf{t}_k$, and one from its corresponding virtual stereo camera viewpoint $\mathbf{E}_k^\text{R} =\mathbf{E}_k \times \mathbf{S} = \mathbf{R}_k|(\mathbf{t}_k+\mathbf{b})$, which represent the reference and the target frames of a perfectly rectified stereo pair, with the latter positioned to the right of the former. This process allows for the generation of countless stereo samples for training deep stereo networks. Additionally, for each viewpoint $\mathbf{E}_k$, we also render a third image from $\mathbf{E}_k^\text{L} = \mathbf{E}_k \times \mathbf{S}^{-1} = \mathbf{R}_k|(\mathbf{t}_k-\mathbf{b})$, which is a second target frame placed on the left of the reference one. This creates a stereo triplet in which the three images are perfectly rectified, as shown in Fig. \ref{fig:losses} (a-c). The importance of this process, particularly in dealing with occlusions, will be discussed in Sec. \ref{sec:loss}. Finally, we extract the disparity $d_\mathbf{r}$ from the rendered depth $z_\mathbf{r}$, which is aligned with the center image of the triplet, and use it to assist in the training of any deep stereo network existing in the literature.

\begin{equation}
    z(\mathbf{r}) = \sum_{i=1}^{N} T_i(1-\text{exp}(-\sigma_i\delta_i))\sigma_i, \quad\quad d(\mathbf{r}) = \frac{b\cdot f}{z(\mathbf{r})}
\end{equation}
with $f$ being the focal length estimated by COLMAP \cite{schoenberger2016sfm}.

\subsection{NeRF-Supervised Training Regime}\label{sec:loss}

Data generated so far is then used to train stereo models. Given a rendered image triplet $(I_l, I_c, I_r)$, we estimate a disparity map $\hat{d}_c$ by feeding the network with $(I_c,I_r)$, which act as the left and right views of a standard stereo pair. Then, we propose an NS loss with two terms.

\textbf{Triplet Photometric Loss.} We exploit image reconstruction to supervise disparity estimation \cite{Godard_CVPR_2017}. Specifically, we backward-warp $I_r$ according to $\hat{d}_c$ and obtain $\hat{I}_c^r$ -- \ie, the reconstructed reference image. Then, we measure the photometric difference between $\hat{I}_c^r$ and ${I}_c$ as:
\begin{equation}
\begin{split}
    \mathcal{L}_\rho(I_{c},\hat{I}_c^r) = \beta\cdot\frac{1- \text{SSIM}(I_{c}, \hat{I_{c}^r})}{2} + (1-\beta)\cdot|I_{c}-\hat{I}_c^r|
\end{split}
\end{equation}
with SSIM being the Structural Similarity Index Measure \cite{WangBSS04}.
Nevertheless, this formulation lacks adequate supervision in occluded regions, such as the left border of the frame or the left of each depth discontinuity, which are not visible in the right image. To overcome this limitation, we employ the third image, $I_l$. By computing $\mathcal{L}_\rho(I_c,\hat{I}_c^l)$, the occlusions will be complementary to those from the previous ones. Thus, to compensate for both, we compute the final, triplet photometric loss defined as the per-pixel minimum \cite{Godard_2019_ICCV} between the two pairwise terms:

\vspace{-0.2cm}
\begin{equation}
\begin{split}
    \mathcal{L}_{\text{3}\rho}(\hat{I_{l}}^c,I_{c}, \hat{I_{r}}^c) = \min \Biggl( \mathcal{L}_\rho(\hat{I_{l}}^c,I_{c}), \mathcal{L}_\rho(I_{c},\hat{I_{r}}^c) \Biggr)
\end{split}
\end{equation}
Fig. \ref{fig:losses} (left)
shows the effect of occlusions when computing $\mathcal{L}_\rho$ between \textit{center-left} (d) and \textit{center-right} (e) pairs with bright colors, whereas they are neglected by $\mathcal{L}_{3\rho}$ (f).
 Finally, untextured regions are discarded by a mask $\mu$ \cite{Godard_2019_ICCV}: 

\begin{equation}
    \mu = [ \min \mathcal{L}_{\text{3}\rho}(\hat{I_{l}}^c,I_{c}, \hat{I_{r}}^c) <  \min \mathcal{L}_{\text{3}\rho}(I_{l},I_{c}, I_{r}) ]
\end{equation}

\textbf{Rendered Disparity Loss.} We further assist the photometric loss by exploiting rendered disparities as:

\begin{equation}
   \mathcal{L}_{disp} = |d_c - \hat{d}_c|
\end{equation}
However, depth maps rendered by NeRF often exhibit artifacts and large errors \cite{deng2022depth}, as shown in Fig. \ref{fig:losses} (g).
To address this issue, we employ a filtering mechanism to preserve only the most reliable pixels. We use Ambient Occlusion (AO) \cite{mueller2022instant} to measure the confidence of $d_c$:

\begin{equation}
    \text{AO} = \sum_{i=1}^N T_i\alpha_i, \quad\quad \alpha_i = 1-\text{exp}(-\sigma_i\delta_i)
\end{equation}
and will use it to filter the disparity loss accordingly. More details are discussed in the \textbf{{supplementary material}}.

\textbf{NeRF-Supervised Loss.} The two terms are summed as:

\begin{equation}
    \begin{split}
    \mathcal{L}_{NS} &= \gamma_{disp}\cdot\eta_{disp}\cdot\mathcal{L}_{disp} \\ &+ \mu \cdot \gamma_{\text{3}\rho}\cdot(1-\eta_{disp})\cdot\mathcal{L}_{\text{3}\rho}    
    \end{split}
\end{equation}
with $\gamma_{disp}, \gamma_{3\rho}$ being weights balancing the impact of photometric and disparity losses, and $\eta_{disp}$ being defined as:

\begin{equation}
    \eta_{disp} = 
    \begin{cases}
    0 & \text{if } \text{AO} < th \\
    \text{AO} & \text{otherwise } \\
    \end{cases}
\end{equation}
according to a threshold $th$ over AO, normalized in $[0,1]$.

%-------------------------------------------------------------------------

\begin{table*}[t]
\centering
    \scalebox{0.6}{
    \renewcommand{\tabcolsep}{15pt}
        \begin{tabular}{lrr|rr|r|r|r|r|rrr|r}
            & \multicolumn{2}{c}{2-views} & \multicolumn{2}{c}{3-views} \\
            \cline{2-13}
            & & & & & & & & KITTI-12 & \multicolumn{3}{c|}{Midd-A} & Midd-21 \\
            & $\mathcal{L}_\rho$ & SGM & $\mathcal{L}_{\text{3}\rho}$ & SGM & $\mathcal{L}_{disp}$ & $>th $ & $\eta_{disp}$ & ($>3$px) & F ($>2$px) & H  ($>2$px) & Q ($>2$px) & ($>2$px) \\
            \cline{2-13}
            \cline{2-13}
            (A) & \checkmark &  &  &  &  &  & & 11.00 & 56.05 & 32.33 & 25.60 & 44.88\\
            (B) & & \checkmark &  &  &  &  & & 6.39 & 22.68 & 17.23 & 18.14 & 23.71\\
            (B') & & \cite{aleotti2020reversing}\checkmark &  &  &  &  & & 5.46 & 19.50 & 15.26 & 17.36 & 21.29 \\
            \cline{2-13}
            (C) & & & \checkmark &  &  & & & 5.11 & 28.33 & 13.57 & 12.09 & 24.38 \\
            (D) & &  &  & \checkmark &  &  & & 5.57 & 19.10 & 13.22 & 14.91 & 20.08 \\
            \cline{2-13}
            (E) & &  &  &  & \checkmark & & & 5.79 & 21.71 & 9.85 & 8.73  &  22.69\\
            (F) & &  &  &  & \checkmark & \checkmark  & & 4.49 & 15.50 & 9.25 & 9.13 & 16.99 \\
            (G) & \checkmark &  &  &  & \checkmark  & \checkmark  & &  4.31 & 16.11 & 9.57 & 9.96 & 16.83\\
            (H) & &  & \checkmark &  & \checkmark  & \checkmark  & & \bfseries 4.21 & 15.31 & 8.86 & 8.41 & 16.26 \\
            (I) & &  & \checkmark &  & \checkmark & \checkmark & \checkmark & 4.31 & \bfseries 14.92 & \bfseries 8.75 & \bfseries 8.28  & \bfseries 14.87 \\
            \cline{2-13}
            \cline{2-13}
            (I') & &  & \checkmark &  & \checkmark & \checkmark & \checkmark & 4.02 & 13.12& 6.91 & 7.18  & 12.87 \\
            \cline{2-13}
        \end{tabular}
    }
    \vspace{-0.3cm}
    \caption{\textbf{Ablation Study -- Loss Components.} Impact of each component in our NS loss. \cite{aleotti2020reversing}$\checkmark$ means using SGM plus \cite{aleotti2020reversing}.} 
    \vspace{-0.3cm}
\label{tab:losses}
\end{table*}

\begin{figure*}
    \centering
    \renewcommand{\tabcolsep}{1pt}
    \begin{tabular}{cccccc}
        \begin{overpic}[height=0.075\textwidth]{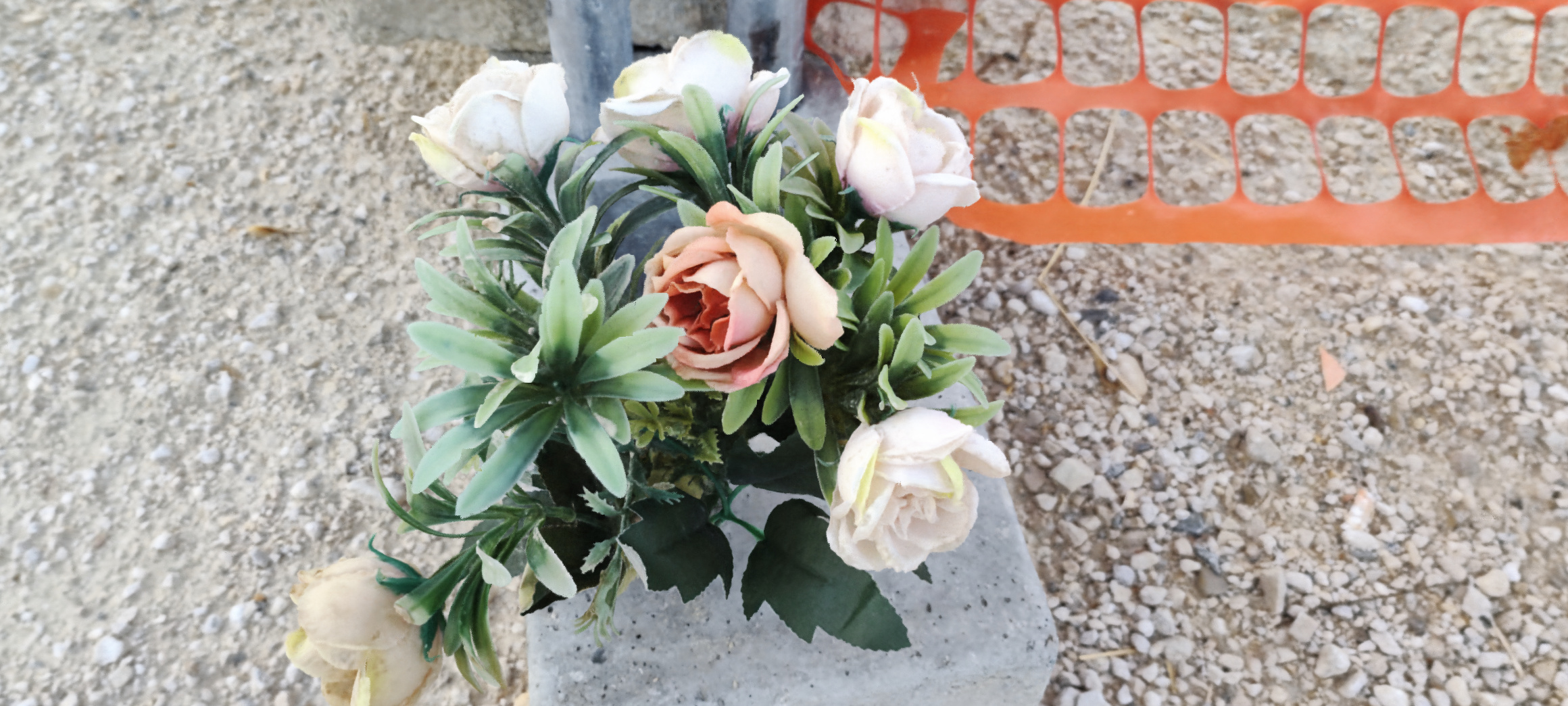}
        \put(0,2){\colorbox{white}{$\displaystyle\textcolor{black}{\text{(a)}}$}}
        \end{overpic}
        &
        \begin{overpic}[height=0.075\textwidth]{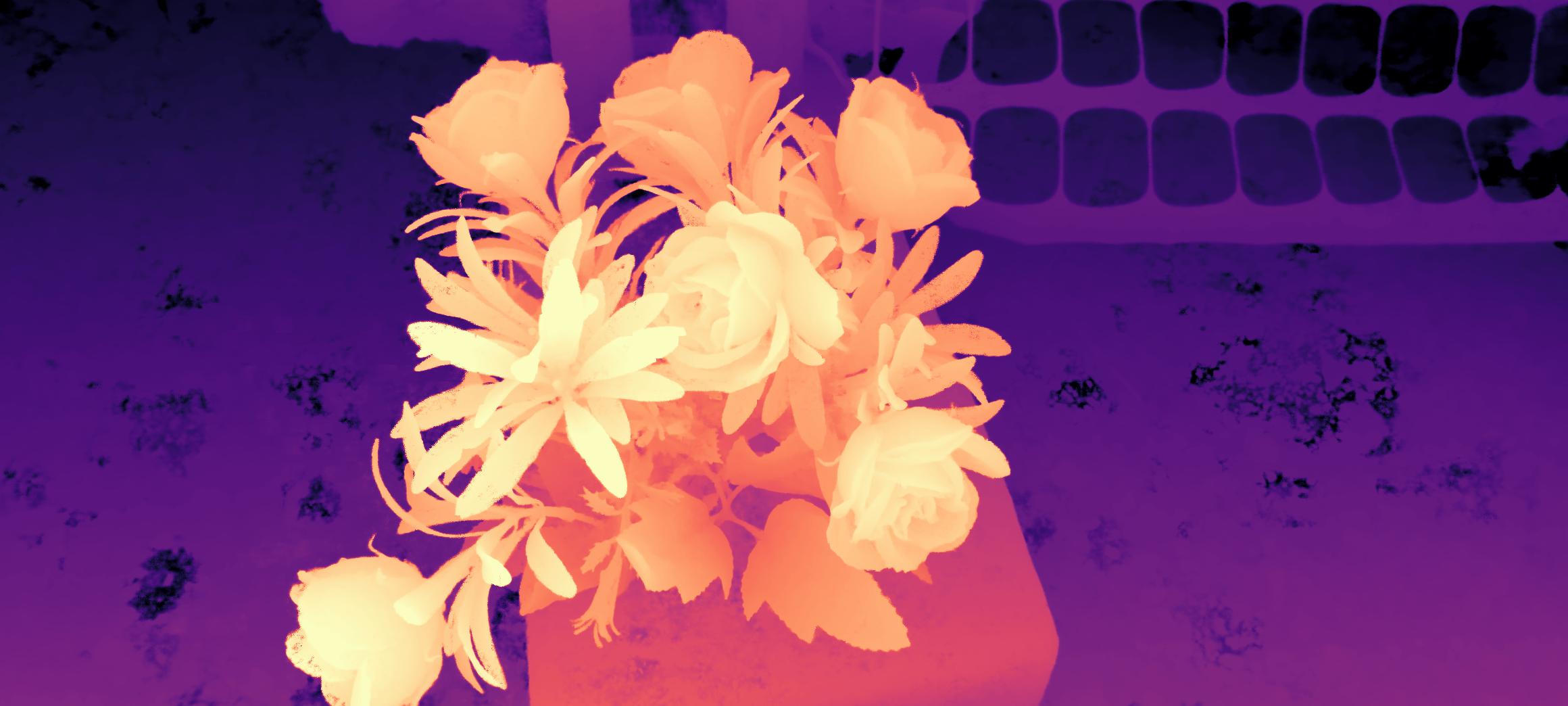}
        \put(0,2){\colorbox{white}{$\displaystyle\textcolor{black}{\text{(b)}}$}}
        \end{overpic}
        & \hspace{0.75cm} &
        \begin{overpic}[height=0.075\textwidth]{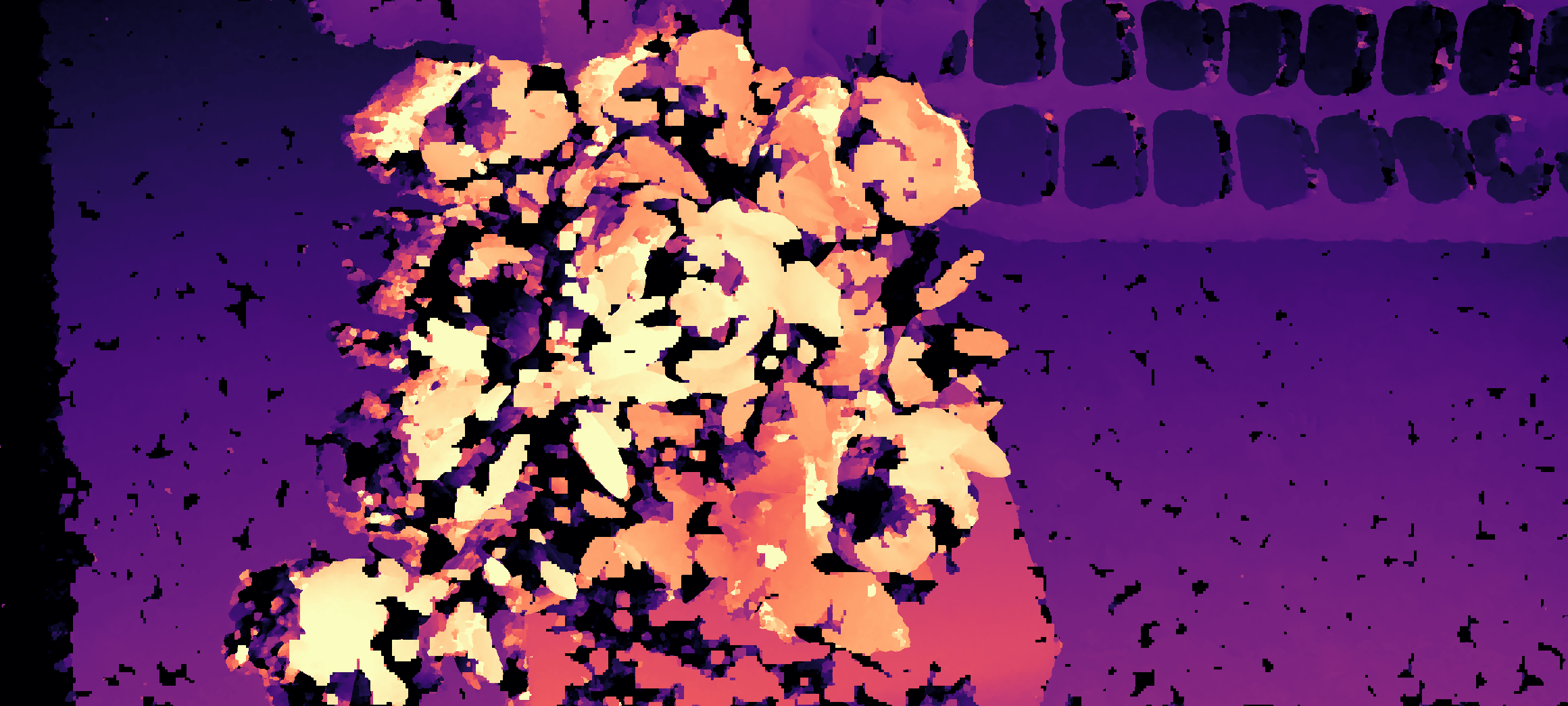}
        \put(0,2){\colorbox{white}{$\displaystyle\textcolor{black}{\text{(c)}}$}}
        \put(23,28){\color{green}\line(2,0){10}}
        \put(23,28){\color{green}\line(0,-2){28}}
        \put(33,28){\color{green}\line(0,-2){28}}
        \put(23,0){\color{green}\line(2,0){10}}
        \put(23,28){\color{green}\line(2,0){10}}
        \put(23,28){\color{green}\line(0,-2){28}}
        \put(33,28){\color{green}\line(0,-2){28}}
        \put(23,0){\color{green}\line(2,0){10}}
        \put(23,28){\color{green}\line(2,0){10}}
        \put(23,28){\color{green}\line(0,-2){28}}
        \put(33,28){\color{green}\line(0,-2){28}}
        \put(23,0){\color{green}\line(2,0){10}}
        \put(23,28){\color{green}\line(2,0){10}}
        \put(23,28){\color{green}\line(0,-2){28}}
        \put(33,28){\color{green}\line(0,-2){28}}
        \put(23,0){\color{green}\line(2,0){10}}
        \end{overpic} 
        \includegraphics[trim=13cm 0 39cm 10cm,, clip,height=0.075\textwidth]{images/0074/raft_bino.png}
        &
        \begin{overpic}[height=0.075\textwidth]{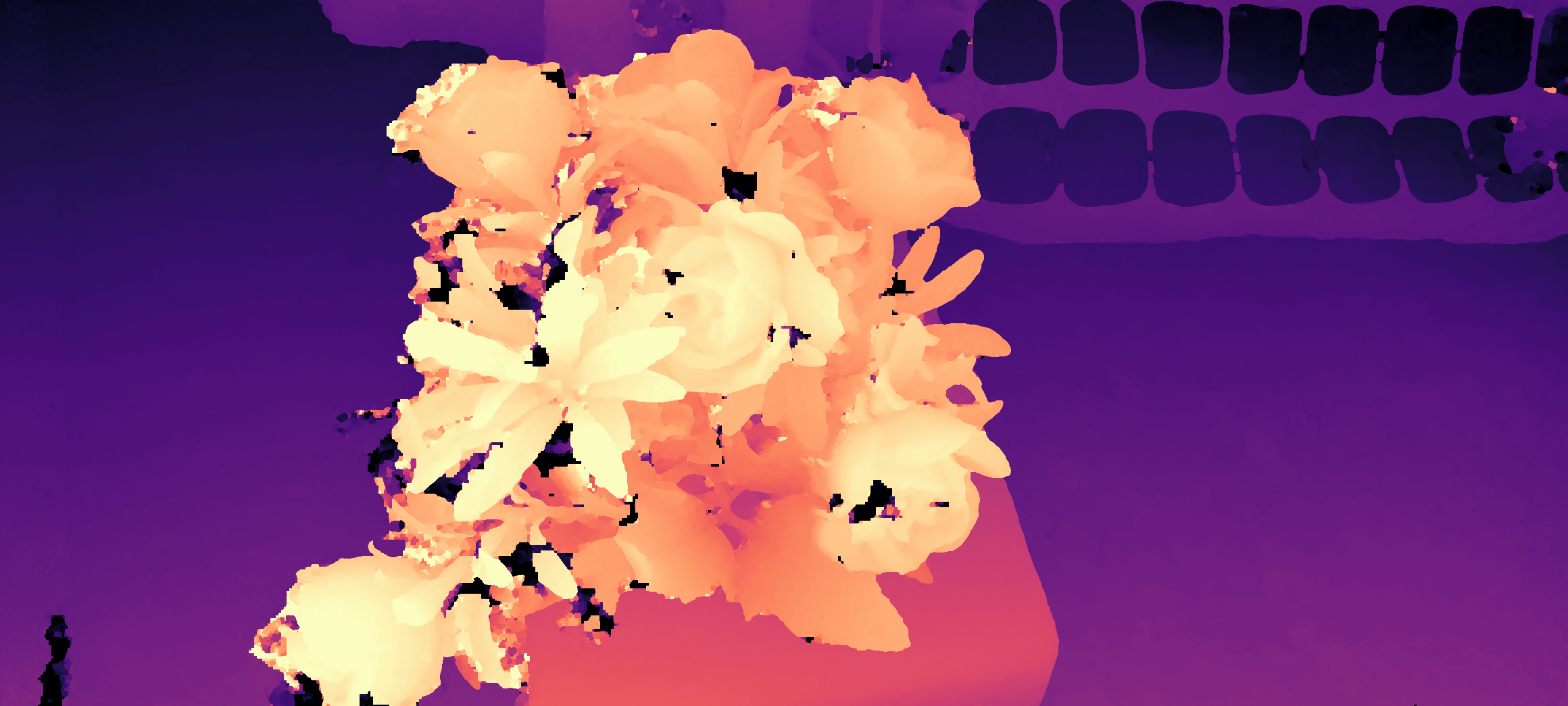}
        \put(0,2){\colorbox{white}{$\displaystyle\textcolor{black}{\text{(d)}}$}}
        \put(23,28){\color{green}\line(2,0){10}}
        \put(23,28){\color{green}\line(0,-2){28}}
        \put(33,28){\color{green}\line(0,-2){28}}
        \put(23,0){\color{green}\line(2,0){10}}
        \put(23,28){\color{green}\line(2,0){10}}
        \put(23,28){\color{green}\line(0,-2){28}}
        \put(33,28){\color{green}\line(0,-2){28}}
        \put(23,0){\color{green}\line(2,0){10}}
        \put(23,28){\color{green}\line(2,0){10}}
        \put(23,28){\color{green}\line(0,-2){28}}
        \put(33,28){\color{green}\line(0,-2){28}}
        \put(23,0){\color{green}\line(2,0){10}}
        \put(23,28){\color{green}\line(2,0){10}}
        \put(23,28){\color{green}\line(0,-2){28}}
        \put(33,28){\color{green}\line(0,-2){28}}
        \put(23,0){\color{green}\line(2,0){10}}
        \end{overpic} 
        \includegraphics[trim=13cm 0 39cm 10cm,, clip,height=0.075\textwidth]{images/0074/raft_trino.png}
        &
        \begin{overpic}[height=0.075\textwidth]{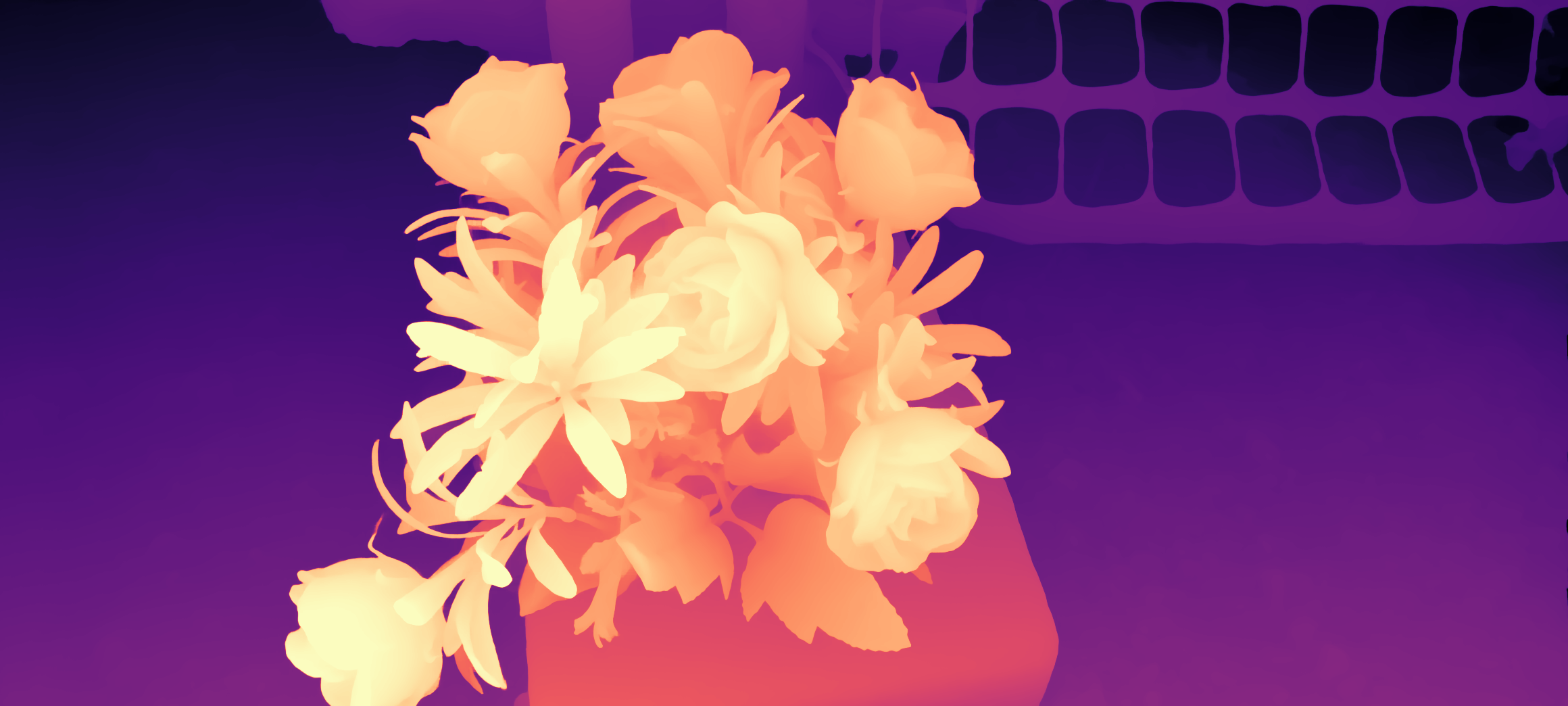}
        \put(0,2){\colorbox{white}{$\displaystyle\textcolor{black}{\text{(e)}}$}}
        \put(23,28){\color{green}\line(2,0){10}}
        \put(23,28){\color{green}\line(0,-2){28}}
        \put(33,28){\color{green}\line(0,-2){28}}
        \put(23,0){\color{green}\line(2,0){10}}
        \put(23,28){\color{green}\line(2,0){10}}
        \put(23,28){\color{green}\line(0,-2){28}}
        \put(33,28){\color{green}\line(0,-2){28}}
        \put(23,0){\color{green}\line(2,0){10}}
        \put(23,28){\color{green}\line(2,0){10}}
        \put(23,28){\color{green}\line(0,-2){28}}
        \put(33,28){\color{green}\line(0,-2){28}}
        \put(23,0){\color{green}\line(2,0){10}}
        \put(23,28){\color{green}\line(2,0){10}}
        \put(23,28){\color{green}\line(0,-2){28}}
        \put(33,28){\color{green}\line(0,-2){28}}
        \put(23,0){\color{green}\line(2,0){10}}
        \end{overpic} 
        \includegraphics[trim=13cm 0 39cm 10cm,, clip,height=0.075\textwidth]{images/0074/raft_ours.png}
        \\
    \end{tabular}
    \vspace{-0.3cm} 
    \caption{\textbf{Effect of Training Losses.} On the left: (a) center image and (b) corresponding disparity rendered by NeRF. On the right: disparity maps (and zoom-in) by RAFT-Stereo trained with (c) center-right or (d) triplet photometric loss, and (e) our full NeRF-Supervised loss.}
    \vspace{-0.3cm}
    \label{fig:losses_ablation}
\end{figure*}

\section{Experimental Results}

We introduce our experiments, first describing implementation details, datasets and, then, discussing our results.

\subsection{Implementation Details}

All experiments are conducted on a single 3090 NVIDIA GPU (more details in the {\textbf{supplementary material}}).

\textbf{Training Data Generation.} We collect a total of 270 high-resolution scenes in both indoor and outdoor environments using standard camera-equipped smartphones. For each scene, we focus on a/some specific object(s) and acquire 100 images from different viewpoints, ensuring that the scenery is completely static. The acquisition protocol involves a set of either front-facing or $\ang{360}$ views.
We use Instant-NGP \cite{mueller2022instant} as the NeRF engine in our pipeline and train it for 50K steps. Running COLMAP and training Instant-NGP takes $\sim25$ minutes per scene, with the collected images having a resolution of $\sim8$Mpx. Afterwards, we generate data with three virtual baselines of $b=0.5, 0.3$ and $0.1$ \textit{units} at different resolutions.
We render a disparity map and a triplet from any image used to train Instant-NGP, aligning the center view to the original viewpoint. This results in a total of 65,148 triplets for training. Although more triplets could have been rendered ( \ie, using additional random viewpoints and baselines), these are sufficient to achieve outstanding results.

\textbf{Deep Stereo Training.} We adopt RAFT-Stereo \cite{lipson2021raft} as the main architecture over which we build our evaluation due to its accuracy and fast convergence. Yet, we also consider PSMNet \cite{chang2018psmnet} and CFNet \cite{shen2021cfnet} to evaluate the effectiveness of our proposal on widely used stereo backbones. We train all models on our dataset with batch size of 2 and a crop size of $384 \times 768$. We run 200k training steps for RAFT-Stereo and 250k for PSMNet and CFNet.  
For ablation experiments, we run 100k iterations following \cite{lipson2021raft}.
All the networks are trained from scratch without any pre-training on synthetic datasets. The augmentation procedure described in \cite{lipson2021raft} is used for training. For PSMNet and CFNet, we set $d_{max}$ to 256 and disabled ImageNet normalization. We use learning rate schedules and optimizers as in \cite{lipson2021raft,chang2018psmnet,shen2021cfnet}. In our experiments, we fix $\beta=0.85$, $th=0.5$, $\gamma_{\text{3}\rho}=0.1$ and $\gamma_{disp}=1$.

\subsection{Evaluation Datasets \& Protocol}

We use the KITTI \cite{Geiger2012CVPR}, Middlebury \cite{scharstein2014high} and ETH3D \cite{schoeps2017cvpr} datasets with publicly available ground-truth for evaluation. Specifically, we define validation and testing splits. 

\textbf{Validation}: 194 stereo images from KITTI 2012, 13 \textit{Additional} images from the training set of Middlebury v3 (\textit{Midd-A}) at Full, Half and Quarter resolutions (F, H, Q), and the Middlebury 2021 (\textit{Midd-21}) dataset. On this split, we run ablation studies and direct comparisons with MfS \cite{watson2020stereo}.

\textbf{Testing}: 200 stereo images from KITTI 2015, 15 stereo pairs from Middlebury v3 training set (\textit{Midd-T}) and 27 pairs from ETH3D. On them, we compare with existing methods that perform zero-shot generalization \cite{zhang2019domaininvariant,chuah2022itsa,zhang2022revisiting,liu2022graftnet}.

\textbf{Evaluation Metrics.} During evaluation, we compute the percentage of pixels having a disparity error greater than a given threshold $\tau$ with respect to the ground-truth. Specifically, we fix $\tau=3$ for KITTI, $\tau=2$ for Middlebury, $\tau=1$ for ETH3D, following the common protocol in the stereo matching field. Unless stated otherwise, we evaluate the computed disparity maps considering both occluded as well as non-occluded regions with valid ground-truth disparity. 

\begin{table}[t]
\centering
    \scalebox{0.6}{
    \renewcommand{\tabcolsep}{10pt}
        \begin{tabular}{ccc|r|rrr|r}
            \hline
            \multicolumn{3}{c|}{Baseline} & KITTI-12 & \multicolumn{3}{c|}{Midd-A} & Midd- 21 \\
            0.5 & 0.3 & 0.1 & ($>3$px) & F ($>2$px) & H ($>2$px) & Q ($>2$px) &  ($>2$px)\\
            \hline
            \checkmark &  - & - & 3.97 & 18.71 & 10.55 & 12.09 & 16.70\\
            \checkmark &  \checkmark & - & \bfseries 3.92 & 16.77 & 9.66 & 10.62 & 16.96\\
            \checkmark &  \checkmark & \checkmark & 4.31 & \bfseries 14.92 & \bfseries 8.75 & \bfseries 8.28 & \bfseries 14.87\\
            \hline
        \end{tabular}
    }
    \vspace{-0.3cm}
\caption{\textbf{Ablation Study -- Impact of Baselines.} We render triplets with \textit{large}, \textit{medium} or \textit{small} baselines -- 0.5, 0.3, 0.1 \textit{units}.}
\vspace{-0.3cm}
\label{tab:baseline}
\end{table}

\begin{figure}[t]
\renewcommand{\tabcolsep}{1pt}
\begin{tabular}{cc}
\includegraphics[width=0.24\textwidth]{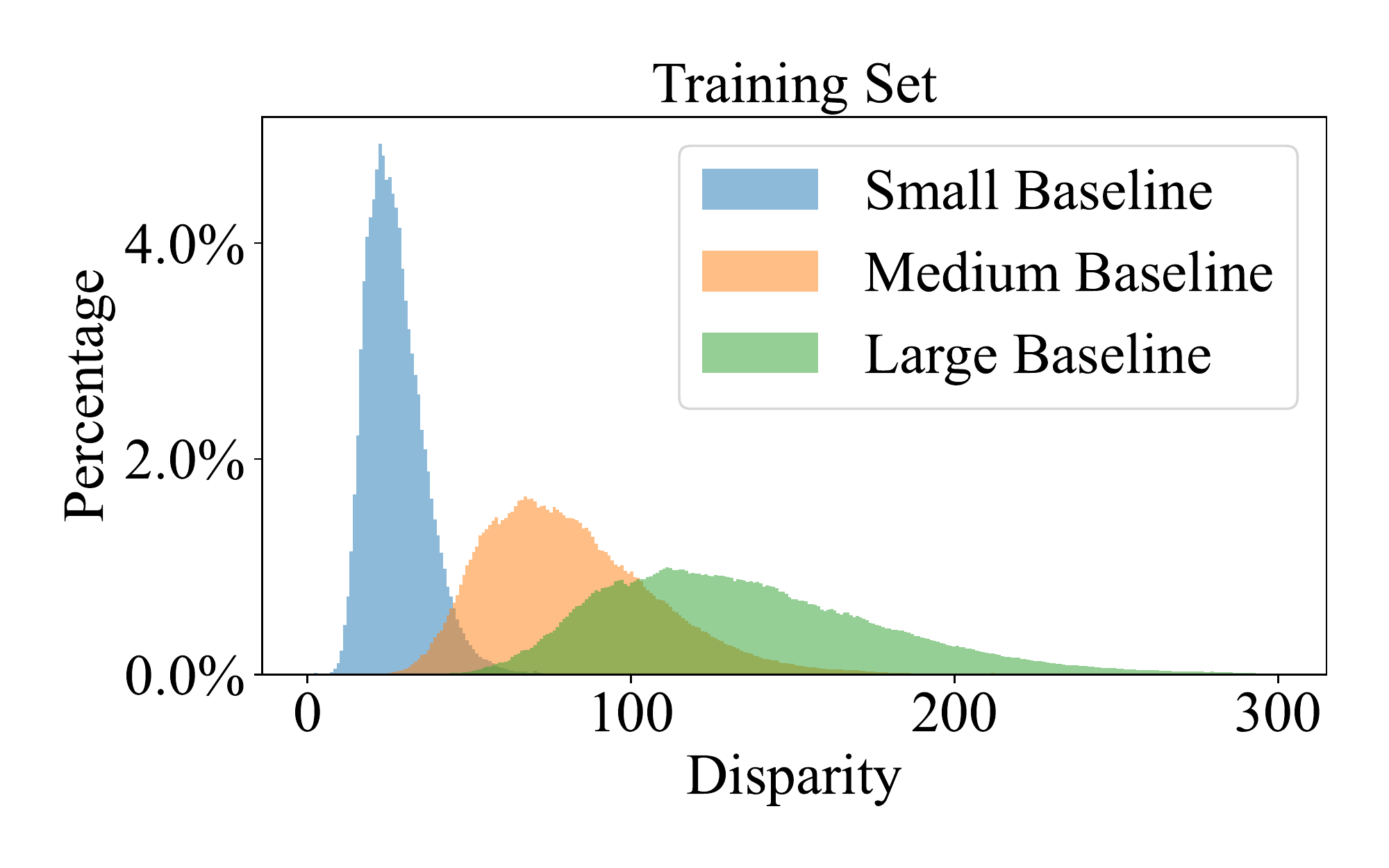} & \includegraphics[width=0.24\textwidth]{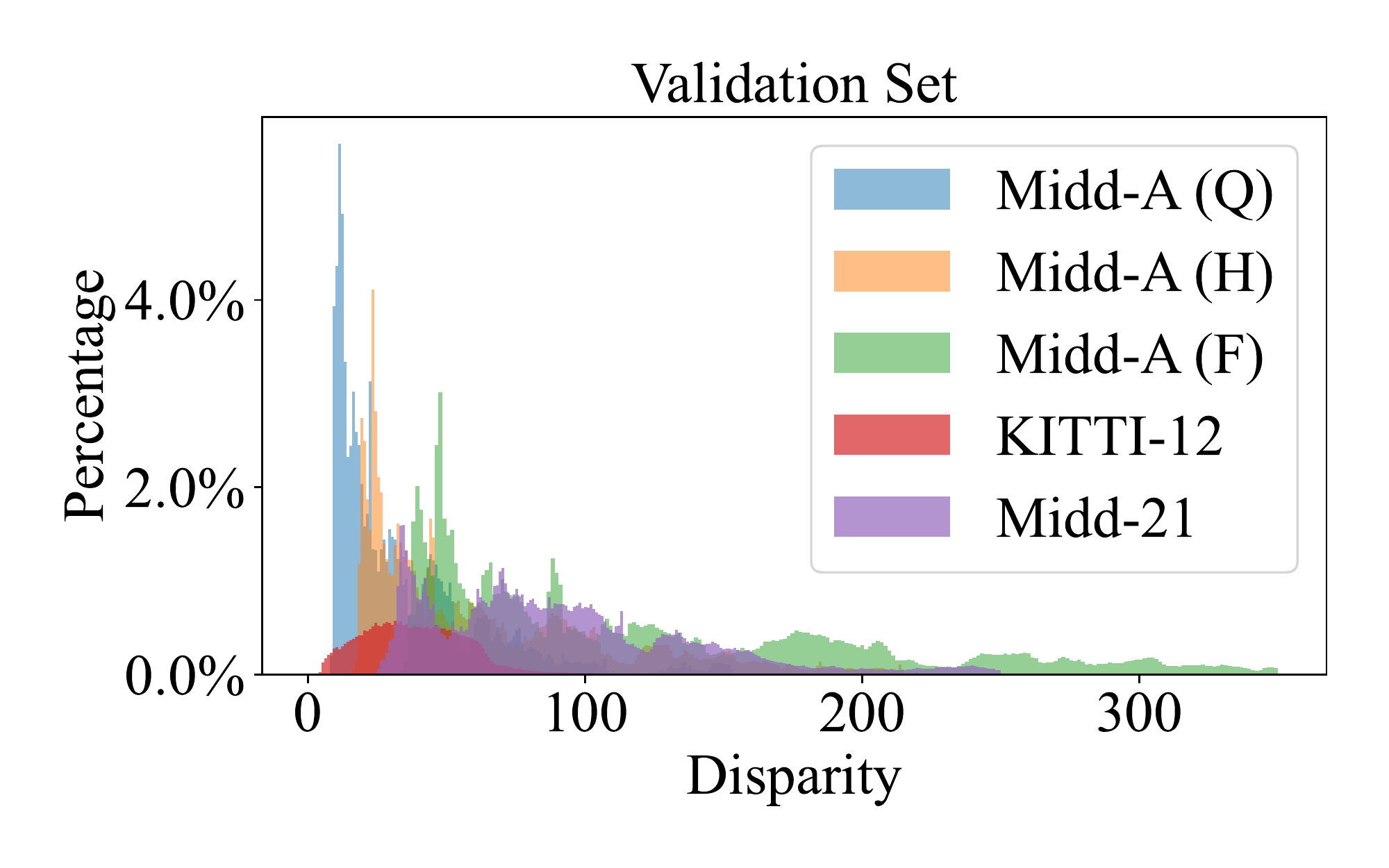}\\
\end{tabular}
\vspace{-0.6cm}
\caption{\textbf{Disparity Distributions.} On the left: our rendered dataset. On the right: datasets from the validation split.}
\vspace{-0.3cm}
\label{fig:distributions}
\end{figure}

\subsection{Ablation Study}

We ablate our NS loss and its impact on the training process, as well as other properties of our rendered dataset. 

\textbf{Loss Analysis.} Tab. \ref{tab:losses}  shows the results of various instances of RAFT-Stereo trained using different variations of our NS loss. 
We start by motivating the use of $\mathcal{L}_{\rho}$: by training the model using a conventional self-supervised loss on stereo pairs (A) leads to poor performance. This is mainly due to occlusions, for which no supervision can be provided.  By using proxy-labels obtained through SGM \cite{hirschmuller2007stereo} and filtering them with a n\"aive left-right consistency check to remove outliers at occlusions (B), the error rates are halved. The same labels processed by \cite{aleotti2020reversing} further improve the results significantly (B'). However, by exploiting the triplets peculiar to our dataset, $\mathcal{L}_{3\rho}$ alone (C) outperforms both (A) and (B), thanks to the stronger self-supervision recovered at occlusions. Interestingly, labels extracted by \cite{aleotti2020reversing} (B') still produce better performance on the high-resolution datasets Midd-A (F) and Midd-21, despite being outperformed by SGM labels over triplets (D).

\begin{table}[t]
\centering
    \scalebox{0.58}{
    \renewcommand{\tabcolsep}{10pt}
        \begin{tabular}{rr|c|ccc|c}
            \hline
            \multicolumn{2}{c|}{Resolution} & KITTI-12 & \multicolumn{3}{c|}{Midd-A} & Midd- 21 \\
            $\sim2$Mpx & $\sim0.5$Mpx & ($>3$px) & F ($>2$px) & H ($>2$px) & Q ($>2$px) &  ($>2$px)\\
            \hline
            \checkmark & - & 5.34 & 14.77 & 11.56 & 12.32 & 15.53\\
            - & \checkmark & \textbf{4.31}  & 14.92 & \textbf{8.75} & \textbf{8.28} & \textbf{14.87}\\
            \checkmark & \checkmark & 4.42  & \textbf{13.92} & 9.12  & 9.66 & 15.88\\
            \hline
        \end{tabular}
    }
    \vspace{-0.3cm}
    \caption{\textbf{Ablation Study -- Impact of Rendering Resolution.} We render images at both half and quarter of the native resolution.}
    \vspace{-0.3cm}
    \label{tab:resolution}
\end{table}

\begin{table}[t]
\centering
    \scalebox{0.6}{
    \renewcommand{\tabcolsep}{10pt}
        \begin{tabular}{rrr|c|ccc|c}
            \hline
            \multicolumn{3}{c|}{\# Scenes} & KITTI-12 & \multicolumn{3}{c|}{Midd-A} & Midd- 21 \\
            65 & 135 & 270 & ($>3$px) & F ($>2$px) & H ($>2$px) & Q ($>2$px) &  ($>2$px)\\
            \hline
            \checkmark & - & - & 3.98 & 18.23 & 11.07 & 11.30 & 17.44\\
             - & \checkmark & - & \textbf{3.87} & 15.82 & 9.69 & 10.36 & 16.51 \\
            - & - & \checkmark & 4.31  & \textbf{14.92} & \textbf{8.75} & \textbf{8.28} & \textbf{14.87}\\
            \hline
        \end{tabular}
    }
    \vspace{-0.3cm}
    \caption{\textbf{Ablation Study -- Number of Collected Scenes.} We render images from different amounts of collected scenes. }
    \vspace{-0.3cm}
    \label{tab:scenes}
\end{table}

\begin{table*}[t]
\centering
    \scalebox{0.58}{
    \begin{tabular}{cc}
        \begin{tabular}{c}
             \\ \\ (A) \\ (B) \\ (C) \\ (D) \\ (E) \\ (F) \\ (G) \\ (H) \\ (I) \\
        \end{tabular}\kern-1.6em & 
         \renewcommand{\tabcolsep}{18pt}
        \begin{tabular}{l|l|l|l|c| r| r r r| r}
            \hline
            \multicolumn{5}{c|}{Configuration} & {KITTI-12} & \multicolumn{3}{c|}{Midd-A} & {Midd-21} \\
            \multicolumn{1}{l}{Model} & \multicolumn{1}{l}{Stereo Network} & \multicolumn{1}{l}{Dataset} & \multicolumn{1}{l}{\# Images} & \multicolumn{1}{l|}{Pre-Train.} & {($>3$px)} & F {($>2$px)} & H {($>2$px)} & Q {($>2$px)} & {($>2$px)} \\
             \hline 
             MfS\cite{watson2020stereo} + MidAs & PSMNet  & MfS & 535K & $\sim$2M & 4.70 & 25.61 & 17.98 & 14.78 & 27.55 \\
             MfS\cite{watson2020stereo} + MidAs & PSMNet & Ours & 65K & $\sim$2M & 5.02 & 28.72 & 20.66 & 16.92 & 26.40 \\
             \rowcolor{cream}
             NS (\textbf{Ours}) & PSMNet & Ours & 65K & 0 & \bfseries 4.07 & \bfseries 19.84 & \bfseries 13.66 & \bfseries 9.15 & \bfseries 19.08  \\
             \rowcolor{white}
             \hline
             MfS\cite{watson2020stereo} + MidAs  & CFNet & MfS & 535K & $\sim$2M & \bfseries 4.47 & 22.00 & 16.69 & 14.32 & 23.44 \\
             MfS\cite{watson2020stereo} + MidAs  & CFNet & Ours & 65K & $\sim$2M & 4.90 & 24.20 & 19.11 & 16.20 & 23.76\\ 
             \rowcolor{cream}
             NS (\textbf{Ours}) & CFNet & Ours & 65K & 0 & 4.64 & \bfseries 17.55 & \bfseries 12.31 & \bfseries 11.13 & \bfseries 19.73\\
             \rowcolor{white}
             \hline
             MfS\cite{watson2020stereo} + MidAs  & RAFT-Stereo & MfS & 535K & $\sim$2M & 4.45 & 19.79 & 12.67 & 9.63 & 22.26\\
             MfS\cite{watson2020stereo} + MidAs  & RAFT-Stereo & Ours & 65K & $\sim$2M & 4.67 & 24.61 & 17.25 & 14.05 & 24.18 \\
             \rowcolor{cream}
             NS (\textbf{Ours}) & RAFT-Stereo & Ours & 65K & 0 & \bfseries 4.02 & \bfseries 13.12 & \bfseries 6.91 & \bfseries 7.18 & \bfseries 12.87\\
             \hline
        \end{tabular} 
    \end{tabular}
    }
    \vspace{-0.3cm}
    \caption{\textbf{Direct Comparison with MfS \cite{watson2020stereo}}. We report results achieved by networks trained using MfS pipeline -- both with their proposed dataset and ours -- and trained with our NeRF-supervised approach (NS). } 
    \vspace{-0.3cm}
\label{tab:watson}
\end{table*}

\begin{figure*}[t]
    \centering
    \renewcommand{\tabcolsep}{1pt}
    \begin{tabular}{cccccc}
        \includegraphics[width=0.16\textwidth]{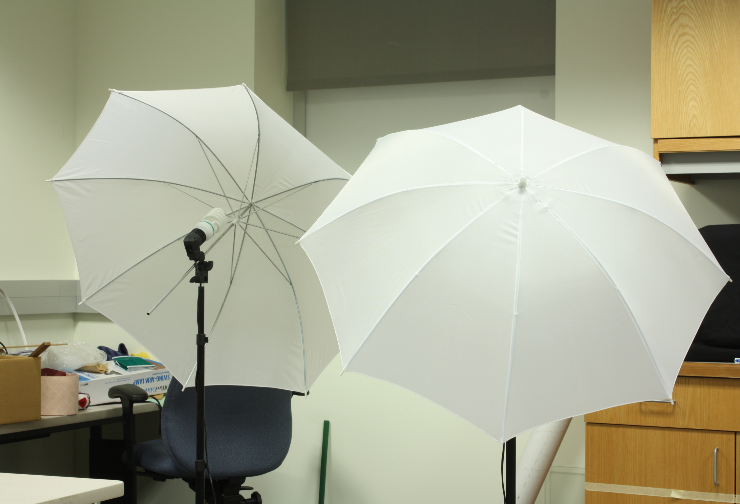} &
        \begin{overpic}[width=0.16\textwidth]{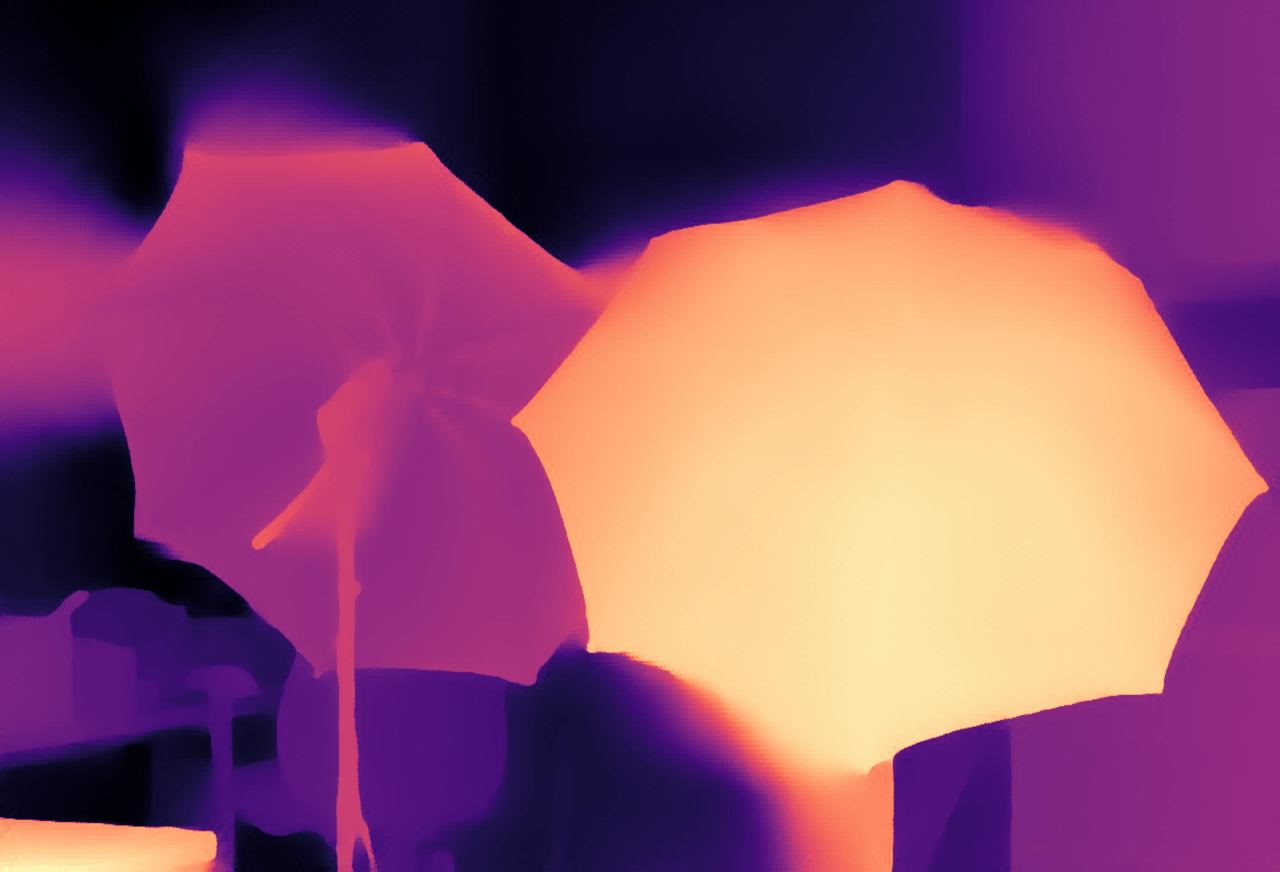}
        %\put(0,30){{$\displaystyle\textcolor{white}{\text{(c)}}$}}
        \end{overpic} & % 35.21
        \includegraphics[width=0.16\textwidth]{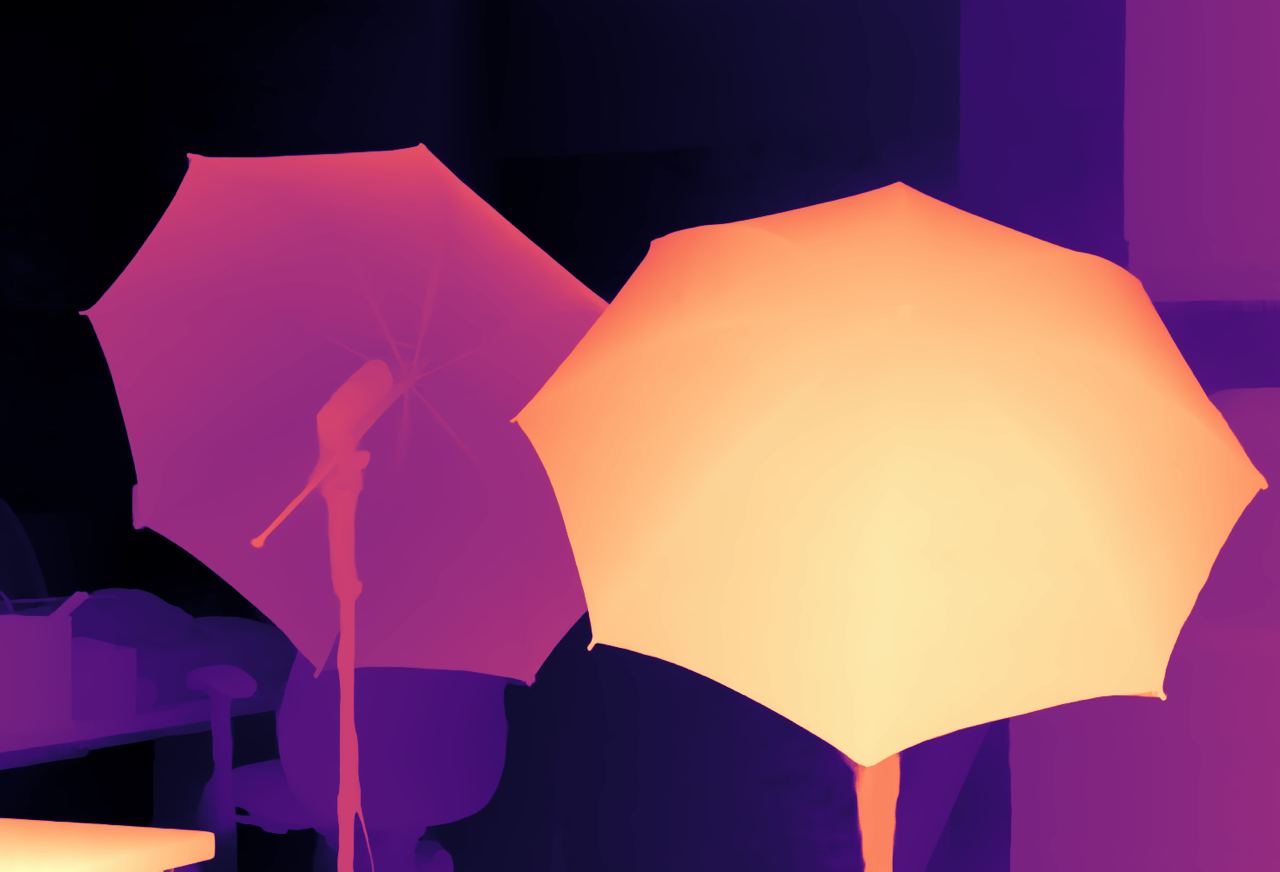} &
        \includegraphics[width=0.16\textwidth]{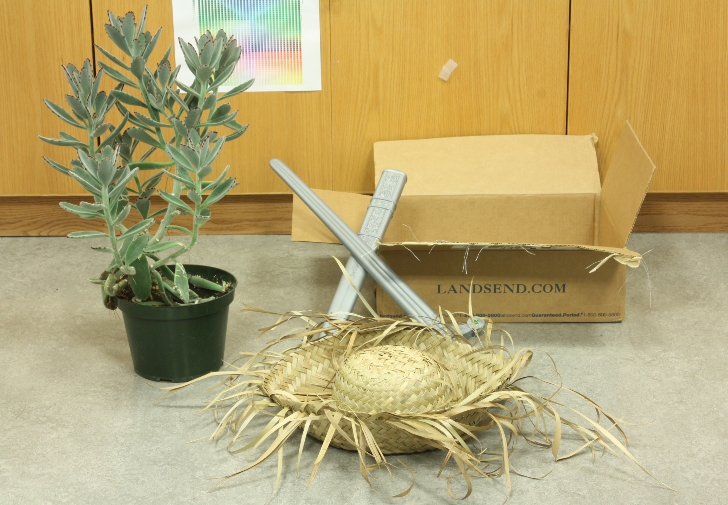} &
        \includegraphics[width=0.16\textwidth]{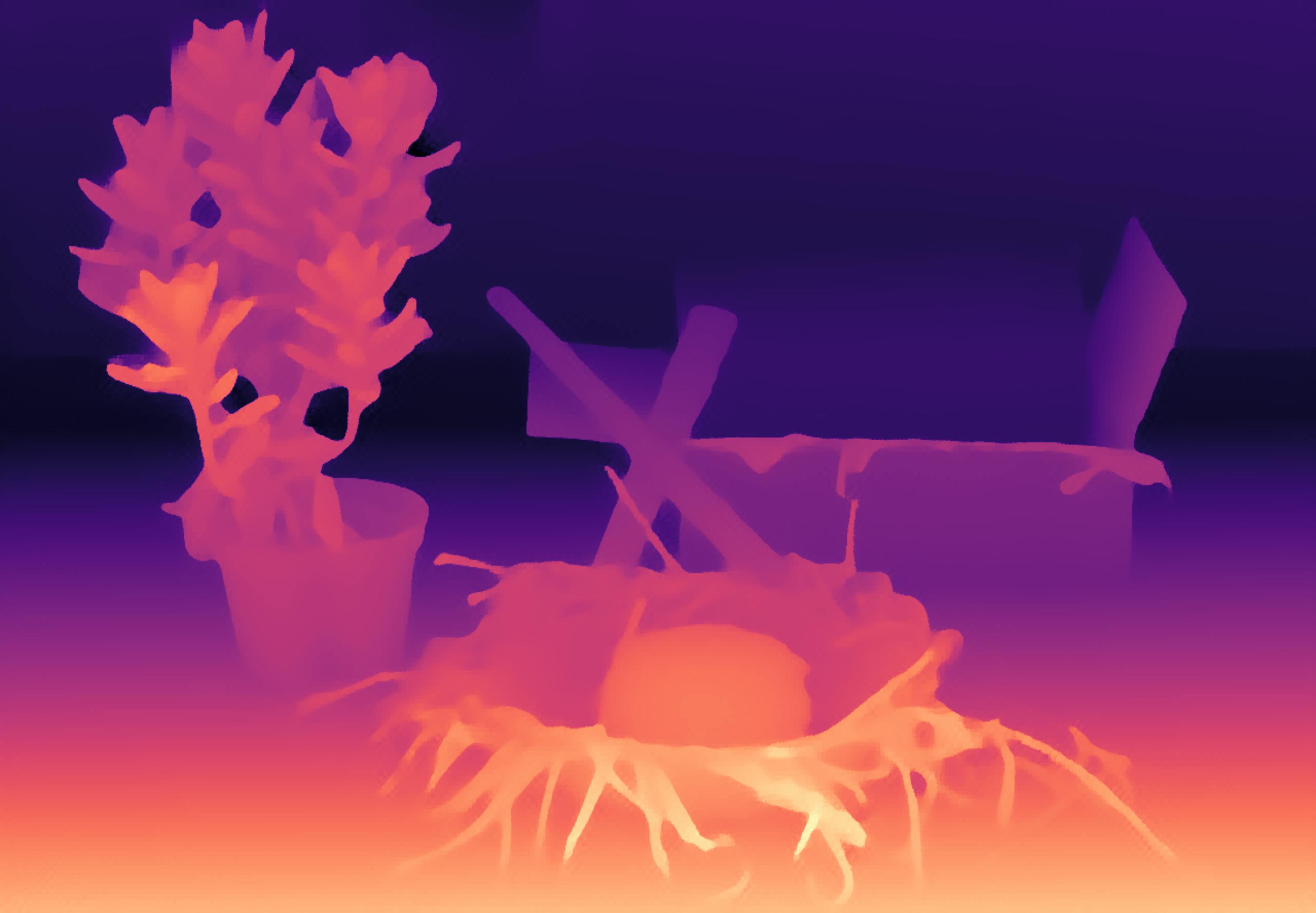} & %14.63
        \includegraphics[width=0.16\textwidth]{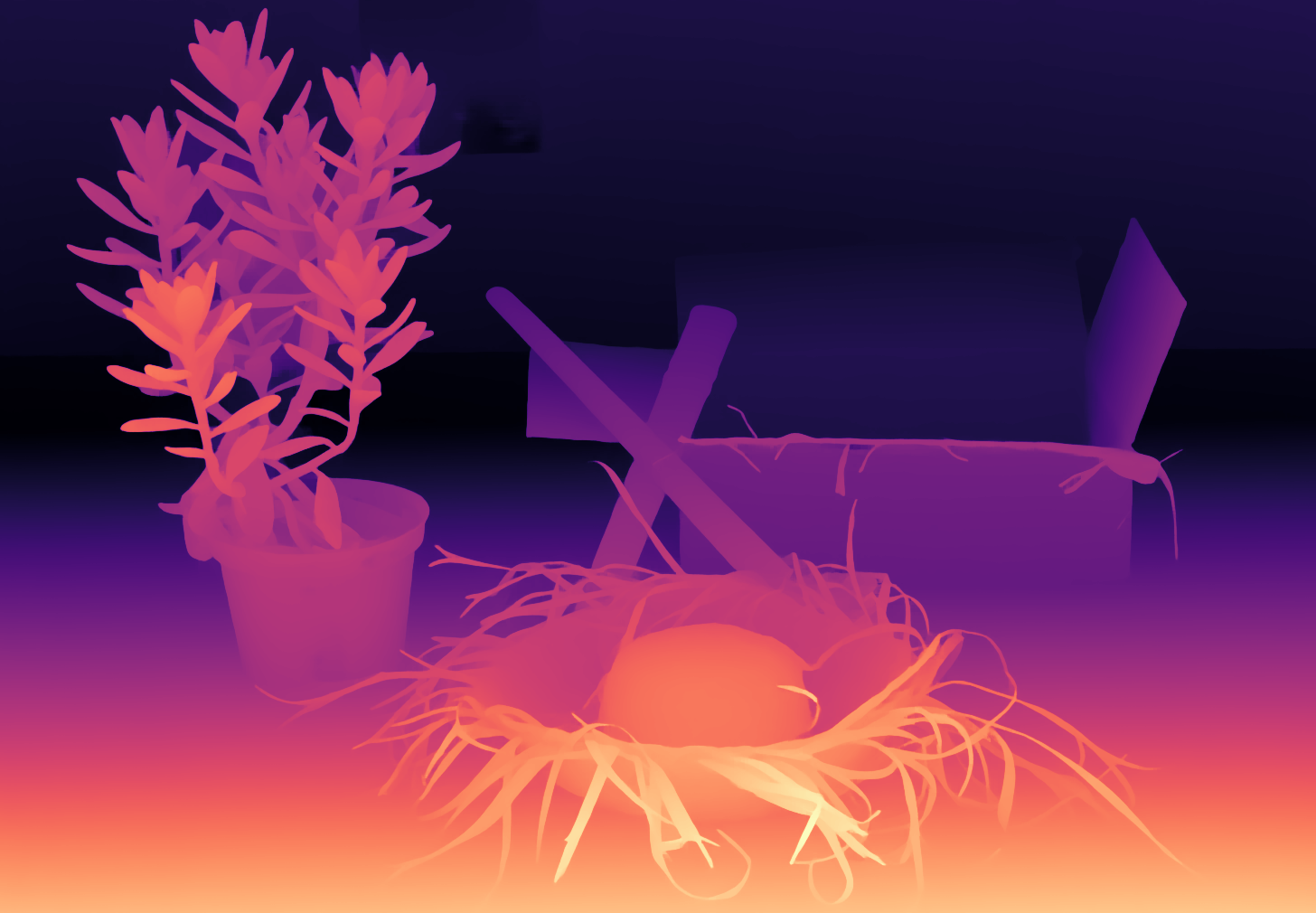}
        \vspace{-0.2cm} \\ 
        \scriptsize ($>2$px) & \scriptsize MfS: 20.47\% & \scriptsize Ours: 6.57\% & & \scriptsize MfS: 8.97\% & \scriptsize Ours: 4.74\%\\
        \includegraphics[trim=0cm 0cm 0cm 0cm,clip,width=0.16\textwidth]{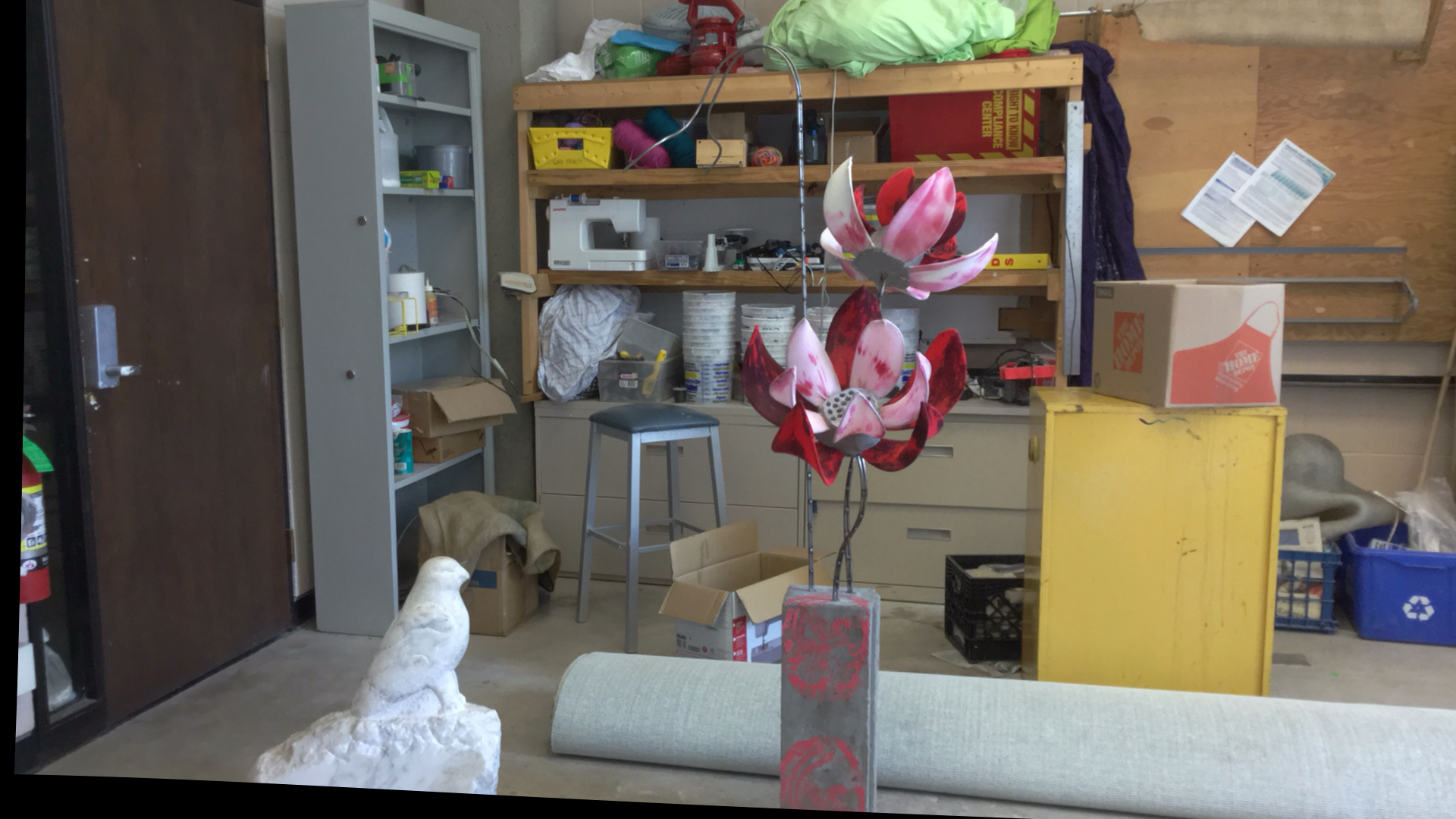} &
        \includegraphics[trim=0cm 0cm 0cm 0cm,clip,width=0.16\textwidth]{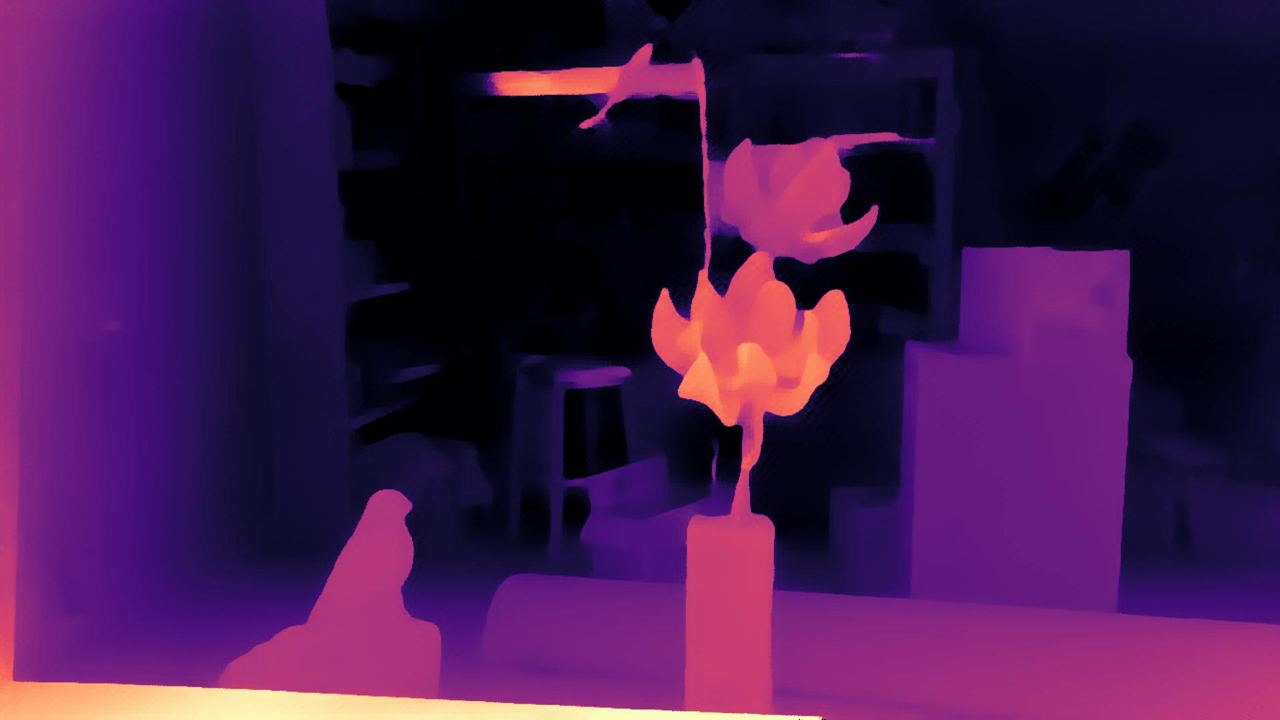} & %14.41
        \includegraphics[trim=0cm 0cm 0cm 0cm,clip,width=0.16\textwidth]{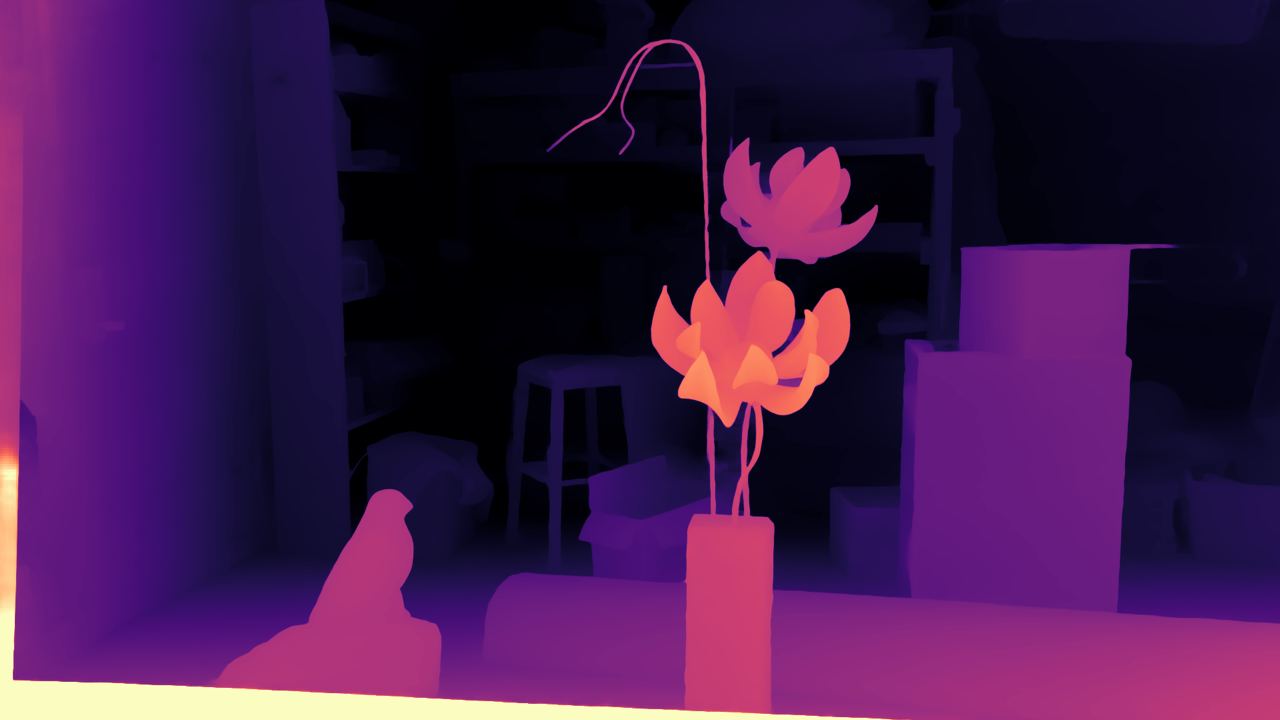} &
        \includegraphics[trim=0cm 0cm 0cm 0cm,clip,width=0.16\textwidth]{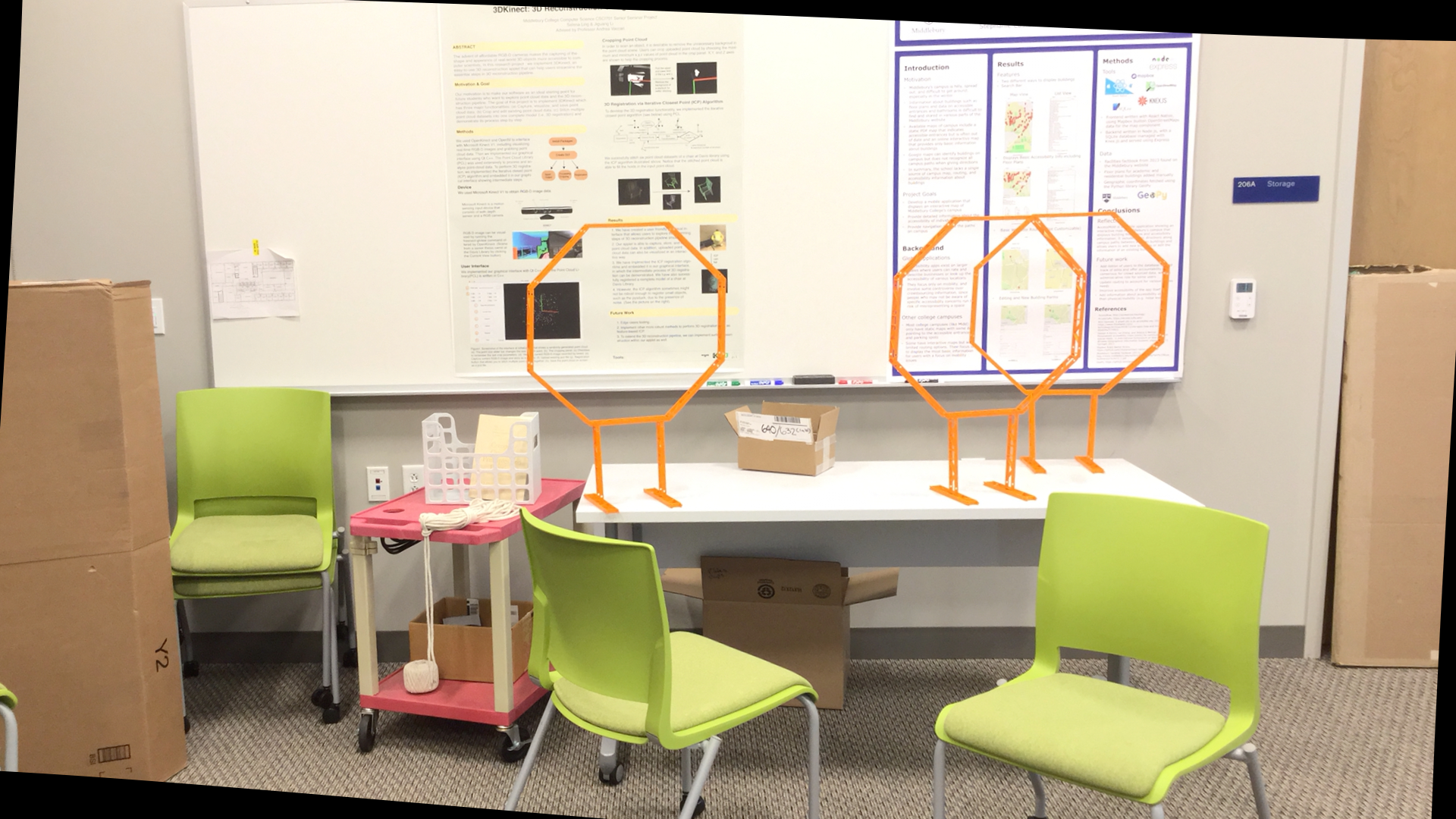} &
        \includegraphics[trim=0cm 0cm 0cm 0cm,clip,width=0.16\textwidth]{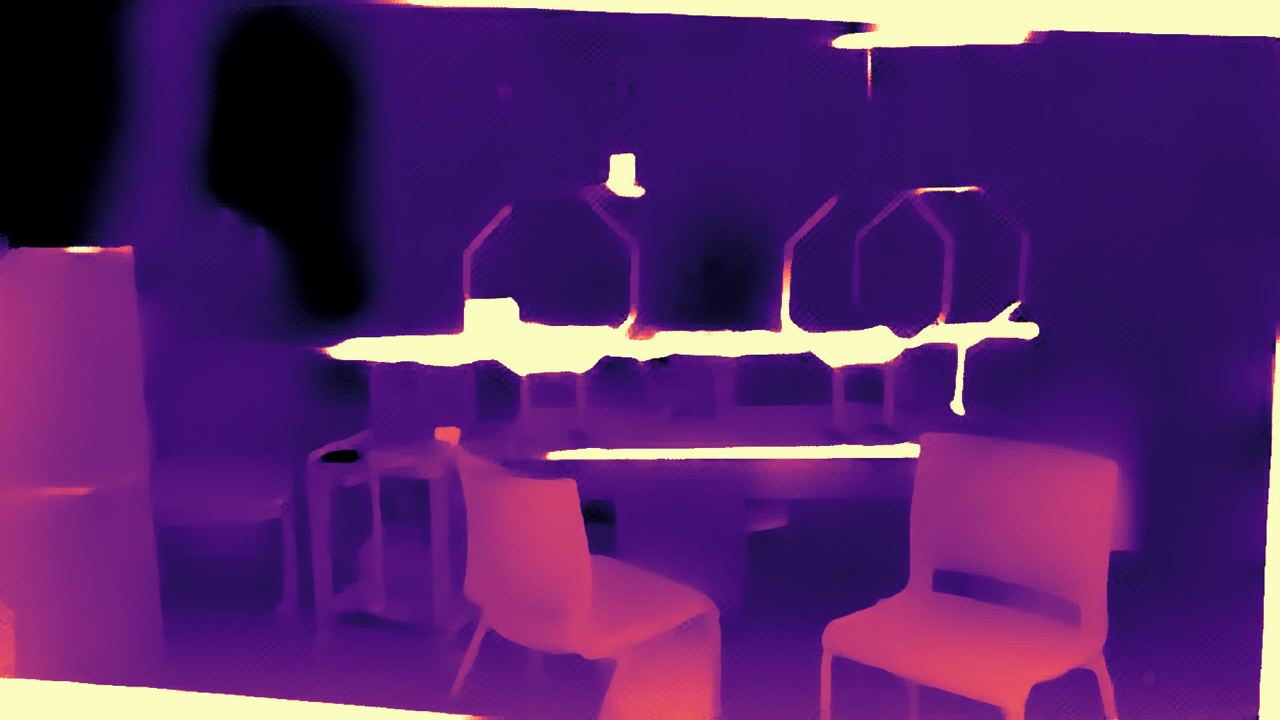} & % 28.45
        \includegraphics[trim=0cm 0cm 0cm 0cm,clip,width=0.16\textwidth]{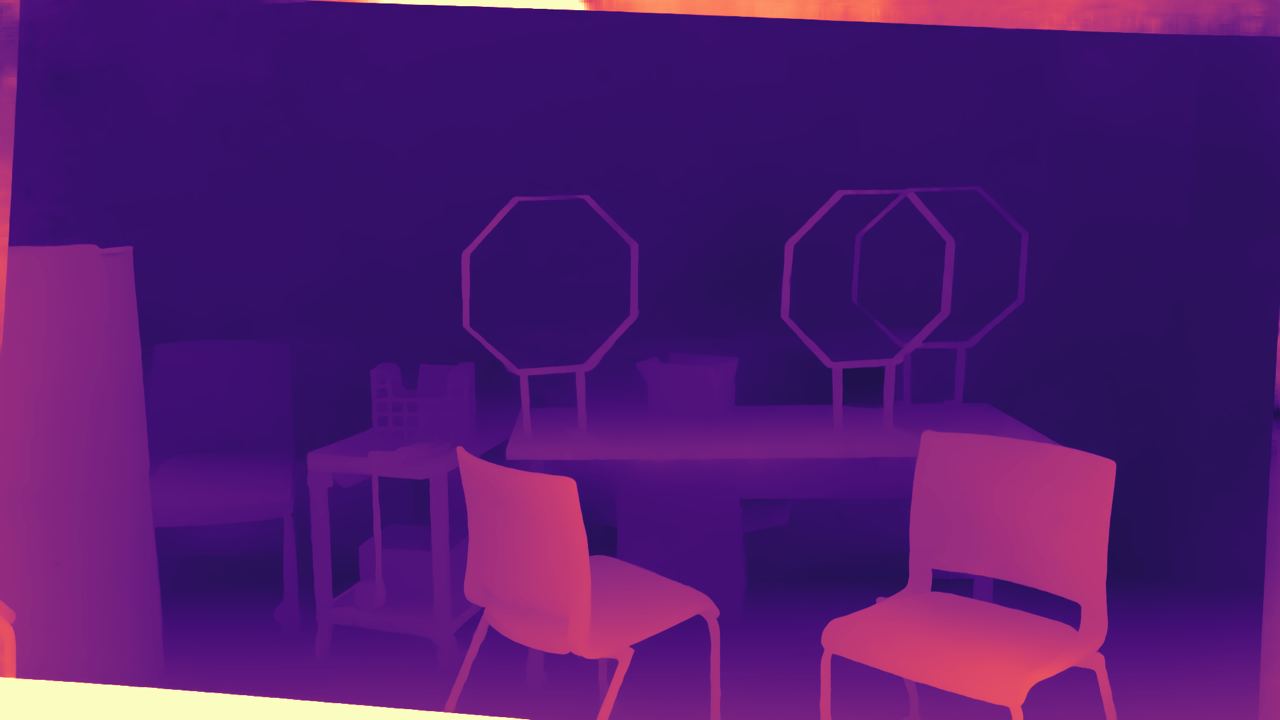}
        \vspace{-0.2cm} \\ 
        \scriptsize ($>2$px) & \scriptsize MfS: 14.41\% & \scriptsize Ours: 6.63\% & & \scriptsize MfS: 24.45\% & \scriptsize Ours: 3.57\%\\
    \end{tabular}
    \vspace{-0.3cm}
    \caption{\textbf{Qualitative Comparison on Midd-A H (top) and Midd-21 (bottom) Datasets.} From left to right: left images and disparity maps by RAFT-Stereo models, respectively trained with MfS or NS. Under each disparity map, the percentage of pixels with error $>2$.}
    \vspace{-0.3cm}
    \label{fig:qualitative}
\end{figure*}

Considering the superiority of proxy labels over photometric losses alone, we then exploit the disparity maps rendered by NeRF to supervise the stereo model (E), resulting in mixed results -- i.e.  better on low-resolution Middlebury but worse on Midd-A (F), Midd-21 and KITTI. Indeed, such a supervision alone results sub-optimal due to the several artefacts shown in Fig. \ref{fig:losses} (g) and recurring in most scenes, making it less effective than $\mathcal{L}_{3\rho}$ on KITTI, Midd-A (F) and Midd-21.
By neglecting the contribution of labels having AO$<th$, results dramatically improve for KITTI and high-resolution datasets, with a minor drop on Midd-A (Q). 
Instead, a major improvement is obtained over all datasets by combining $\mathcal{L}_{disp}$ with $\mathcal{L}_{3\rho}$ (H). The triplet is crucial for this: combining $\mathcal{L}_{disp}$ with $\mathcal{L}_\rho$ is less effective (G).

Finally, our full $\mathcal{L}_{NS}$ loss (I), balancing the two terms according to AO, results the most effective on the validation split. Furthermore, row (I') shows the impact of a longer training schedule (200K steps vs 100K).
Fig. \ref{fig:losses_ablation} qualitatively shows how the estimated disparities by RAFT-Stereo improve dramatically when switching from conventional image loss (c) to $\mathcal{L}_{3\rho}$ (d), although finer details are still missing. $\mathcal{L}_{NS}$ recovers them with unprecedented fidelity (e).

\begin{table*}[t]
\centering
    \scalebox{0.56}{
            \begin{tabular}{cc}
                \begin{tabular}{c}
                 \\ \\ \\ \\ \\ \\ \\ \\ \\ \\ (A) \\ \\ \\ \\ \\ \\ \\ \\ \\ (B) \\ \\ \\ \\ \\ \\ (C) \\ \\ (D) \\
                \end{tabular}\kern-1.6em  
            &
            \renewcommand{\tabcolsep}{20pt}
            \begin{tabular}{lc|rr|rrrrrr|rr}
                \hline
                & & \multicolumn{2}{c|}{KITTI-15} & \multicolumn{6}{c|}{Midd-T} & \multicolumn{2}{c}{ETH3D} \\
                Method & &  \multicolumn{2}{c|}{($>3$px)} & \multicolumn{2}{c}{F ($>2$px) } & \multicolumn{2}{c}{H ($>2$px)} & \multicolumn{2}{c|}{Q ($>2$px)} & \multicolumn{2}{c}{($>1$px)} \\
                \hline
                & & All & Noc & All & Noc & All & Noc & All & Noc & All & Noc \\
                \hline
                Training Set & \multicolumn{10}{c}{SceneFlow with GT} \\
                \hline
                %\rowcolor{LightYellow}            
                \rowcolor{white}
                %\rowcolor{LightPink}
                GANet \cite{zhang2019ga} & & 10.46 & 10.15 & 45.36 & 40.80 & 26.75 & 21.8 & 15.52 & 11.49 & 8.68 & 7.75 \\
                %\rowcolor{LightPink}
                DSMNet \cite{zhang2019domaininvariant} & & 5.50 & 5.19 & 29.95 & 24.79 & 16.88 & 12.03 & 13.75 & 9.44 & 12.52 & 11.62 \\
                %\rowcolor{LightPink}
                CFNet \cite{shen2021cfnet} & & 6.01 & 5.94 & 29.12 & 24.15 & 20.11 & 15.84 & 13.77 & 10.32 & 5.77 & 5.32 \\
                \rowcolor{white}
                MS-GCNet \cite{cai2020matchingspace} & * & 6.21 & - & - & - & - & 18.52 & - & - & - & 8.84 \\
                \cdashline{3-12}
                RAFT-Stereo \cite{lipson2021raft} & * & 5.74 & - & 18.33 & - & 12.59 & - & 9.36 & - & 3.28 & - \\
                %\rowcolor{LightYellow}
                RAFT-Stereo \cite{lipson2021raft} & \textdaggerdbl &  5.45 & 5.21 & \bfseries \silver{18.20} & \bfseries \silver{14.19} & \bfseries \silver{11.19} & \bfseries \silver{8.09} & \bfseries \silver{9.31} & \bfseries \silver{6.56} & \bfseries \gold{2.59} & \bfseries \silver{2.24} \\
                \cdashline{3-12}
                SGM + NDR \cite{aleotti2021neural} & &  5.41 & 5.12 & 27.27 & 21.09 & 17.70 & 13.51 & 11.75 &  7.93 & \bronze{5.20} & \bronze{4.78} \\         
                %\rowcolor{LightPink}
                STTR \cite{li2021revisiting} & &  8.31 & 6.73 & 38.10 & 30.74 & 26.39 & 18.17 & 15.91 & 8.51 & 20.49 & 19.06\\ % MASSIMA RISOLUZIONE MIDDLEBURY: HALF * 0.85
                %\rowcolor{LightPink}
                CEST \cite{guo2022context} & &  7.61 & 6.13 & 27.44 & 19.53 & 19.89 & 11.82 & 14.71 & 7.56 & 10.99 & 9.78\\ % MASSIMA RISOLUZIONE MIDDLEBURY: HALF
                %\rowcolor{LightPink}
                FC-GANet \cite{zhang2022revisiting} & * & 5.3x & - & - & - & - & 10.2x & - & 7.8x & 5.8x & - \\
                ITSA-GWCNet \cite{chuah2022itsa} & &  5.60 & 5.39 & 29.46 & 25.18 & 19.38 & 15.95 & 14.36 & 10.76 & 7.43 & 7.12 \\
                ITSA-CFNet \cite{chuah2022itsa} & &  \bfseries \gold{4.96} & \bfseries \gold{4.76} & 26.38 & 21.41 & 18.01 & 14.00 & 13.32 & 9.73 & 5.40 & 5.14 \\       
                CREStereo \cite{li2022practical} & \textdaggerdbl & 5.79 & 5.40 & 34.78 & 30.52 & 17.57 & 13.87 & 12.88 & 8.85 & 8.98  & 8.14 \\
                \hline
                PSMNet \cite{chang2018psmnet} & \textdaggerdbl & 7.86 & 7.40 & 33.69 & 28.35 & 21.69 & 16.92 & 17.24 & 12.37 & 23.19 & 22.12 \\
                MS-PSMNet \cite{cai2020matchingspace} & * &  7.76 & - & - & - & - & 19.81 & - & - & - & 16.84 \\
                Graft-PSMNet\cite{liu2022graftnet} & &  \bfseries 5.34 & \bfseries 5.02 & \bfseries 25.46 & \bfseries 19.28 & \bfseries 17.81 & \bfseries 13.46 & \bfseries 14.18 & \bfseries 9.21 & 11.42 & 10.69 \\
                FC-PSMNet \cite{zhang2022revisiting} & * & 5.8x & - & - & - & - & 15.1x & - &  9.3x & \bfseries 9.5x & - \\
                ITSA-PSMNet \cite{chuah2022itsa} & &  6.00 & 5.73 & 32.09 & 27.46 & 20.83 & 17.14 & 14.68 & 11.05 & 10.34 & \bfseries 9.77 \\   
                \hline
                \hline
                Training Set & \multicolumn{10}{c}{Real-world data without GT} \\
                \hline 
                Reversing\cite{aleotti2020reversing}-PSMNet & & $[4.09]$ & $[3.88]$ & 38.23 & 30.00 & 26.45 & 20.91 & 20.55 & 15.08 & 9.00 & 8.23 \\
                MfS\cite{watson2020stereo}-PSMNet & & \bronze{5.18} & \bronze{4.91} & 26.42 & 21.38 & 17.56 & 13.45 & 12.07 & 9.09 & \bfseries 8.17 & \bfseries 7.44 \\
                \rowcolor{cream}
                NS-PSMNet (\textbf{Ours}) & & \bfseries \silver{5.05} & \bfseries \silver{4.80} & \bfseries \bronze{20.60} & \bfseries \bronze{15.83} & \bfseries \bronze{12.91} & \bfseries \bronze{9.07} &  \bfseries \bronze{11.03} & \bfseries \bronze{7.15} & 11.69 & 11.00 \\
                \hline
                \rowcolor{cream} 
                NS-RAFT-Stereo (\textbf{Ours}) & & 5.41 & 5.23 & \bfseries \gold{16.45} & \bfseries \gold{12.08} & \bfseries \gold{9.67} & \bfseries \gold{6.42} & \bfseries \gold{8.05} & \bfseries \gold{4.82} & \bfseries \silver{2.94} & \bfseries \gold{2.23} \\
    
                \hline
            \end{tabular}
            \end{tabular}
    }
    \vspace{-0.3cm}
    \caption{\textbf{Zero-Shot Generalization Benchmark.} We test all models using authors' weights. Exceptions: $*$ numbers from the original paper; \textdaggerdbl{} retrained model.
    Best results per macro-block in \textbf{bold}. We also highlight \gold{\textbf{first}}, \silver{\textbf{second}} and \bronze{\textbf{third}} absolute bests. For RAFT-Stereo (in dashed lines) only the best between $*$ and \textdaggerdbl{} is kept for rankings. $[\kern.2em]$ means trained on the same domain, thus ignored for rankings. 
    }\vspace{-0.3cm}
\label{tab:zero}
\end{table*}

\textbf{Impact of Virtual Baselines.} We evaluate the impact of the virtual baseline used to render triplets on the disparity distribution. Tab. \ref{tab:baseline} shows the results of our study on the training effectiveness. Specifically, we render our dataset using a single, large baseline of 0.5 units (21,716 triplets), as well as adding more images obtained with medium and small baselines of 0.3 and 0.1 units. We can observe that using only the large one yields the best results on KITTI, while rendering additional images with the medium baseline results in improvements on Midd-A only. Utilizing all three baselines leads to the best results on Midd-A and Midd-21, with a moderate drop on KITTI. We ascribe this to the disparity distributions generated by using the three baselines, that covers the full range defined by the combined validation sets, as shown in Fig. \ref{fig:distributions}. %Indeed, larger baselines (0.3, 0.5) better cover the stretched distribution observed on KITTI. 

\textbf{Impact of Image Resolution.} We evaluate the impact of image resolution on the training process. Purposely, we render images at approximately 2 and 0.5Mpx out of the original 8Mpx images -- this because, in terms of computational burden, existing stereo networks can rarely deal with them, despite our pipeline would perfectly allow for this. As shown in Tab. \ref{tab:resolution}, the best results are usually obtained by rendering 0.5Mpx images, except when testing at full resolution on Midd-A, for which rendering both higher and lower resolution images provides benefits.

\textbf{Impact of Scenes.} Finally, we show the impact of a larger number of collected scenes on the training process. Tab. \ref{tab:scenes} highlights how the accuracy on the most challenging datasets -- i.e., Middlebury -- increases with it, unsurprisingly. This represents a key strength of our work, enabling anyone to generate their own extensive and scalable training data collections for stereo, resulting in better and better results, thanks to its ease of implementation.

\subsection{Comparison with MfS}

To further evaluate the quality of our rendered data, we compare our approach with MfS \cite{watson2020stereo} -- the most recent method for generating stereo pairs from single images -- by training three different stereo networks. Tab. \ref{tab:watson} collects the outcome of this experiment using PSMNet \cite{chang2018psmnet}, i.e. the baseline model used in \cite{watson2020stereo}, as well as with CFNet \cite{shen2021cfnet} and RAFT-Stereo \cite{lipson2021raft}. Each model is trained on the dataset proposed in \cite{watson2020stereo} (A,D,G), as well as on stereo pairs generated with their technique on our data (B,E,H) or by means of our NS paradigm (C,F,I). We point out that in the first two cases, 2 million labeled images were used to train MidAS \cite{Ranftl2022}, which is the key component of their pipeline. In contrast, our approach does not require any additional data.

First, we note that using the MfS generation method on our data consistently leads to inferior results compared to using theirs -- i.e. (B) vs (A), (E) vs (D), (H) vs (G). This is unsurprising, considering that our images were collected from only 270 scenes, while the original dataset used by MfS includes half a million images from COCO, ADE20K, DIODE, Mapillary, and DiW, which provide a much wider range of scenes and contexts. This excludes the fact that the superior results achieved by NS are a consequence of the quality of collected images solely.
Eventually, any network trained with the NS supervision granted by our data, always outperforms its two counterparts by a large margin -- except with CFNet on KITTI, where we register a 0.17\% drop. This proves that our paradigm is effective with different stereo architectures 
and consistently outperforms MfS without the need for a large training dataset. 
%-- not only RAFT-Stereo -- and consistently outperforms MfS. 

Finally, Fig. \ref{fig:qualitative} shows a comparison between RAFT-Stereo trained with MfS on the dataset adopted in \cite{watson2020stereo}, and its NS counterpart.
The results showcase the much more detailed predictions of the latter, especially in  thin structures, which is of unprecedented quality for methods not trained on ground-truth. More are reported in the \textbf{{supplement}}.

\subsection{Zero-Shot Generalization Benchmark}

We conclude by evaluating stereo networks trained in an NS manner for zero-shot generalization. Table \ref{tab:zero} collects the comparison with several state-of-the-art methods on the benchmark common to the latest works \cite{chuah2022itsa,zhang2022revisiting,liu2022graftnet}.

It is worth mentioning that different papers have often evaluated with different protocols\footnote{We reached this verdict by checking the authors' code and, when not available, through private communications with authors themselves.} on the Middlebury dataset: some compute the metrics only for \textit{Noc} regions \cite{liu2022graftnet,chuah2022itsa}, some limit the set of valid pixels to those having ground-truth disparity lower than 192\cite{liu2022graftnet,chuah2022itsa}, and others compute a weighted average over the dataset, setting challenging images to 0.5 as indicated on the Middlebury website \cite{aleotti2020reversing,chuah2022itsa}. The protocol itself is often not reported in the papers, leading to the accumulation of several inconsistencies throughout the literature and, possibly, drawing biased conclusions. To address this, we re-evaluated any method with available code and weights, both over \textit{All} / \textit{Noc} pixels and considering the entire disparity range, to establish a common protocol from now on. For a few methods whose weights are no longer available, we either took numbers from the original paper ($*$), although they may not be entirely comparable with the others, or retrained them (\textdaggerdbl). 

We defined four main groups of methods: (A) existing stereo models, excluding (B) PSMNet variants, both of which were trained on synthetic data with ground-truth; (C) PSMNet models trained without ground-truth; and (D) our best model. (B) and (C) allow for comparison of several methods pursuing generalization while using a common backbone. We can see that our NS-PSMNet outperforms all PSMNet variants, except on ETH3D (it is worth noting that Reversing-PSMNet \cite{aleotti2020reversing} was trained on raw KITTI \cite{Godard_CVPR_2017}, which gives it a significant advantage). Among the methods in group (A), only RAFT-Stereo outperforms NS-PSMNet on Middlebury, but performs worse on KITTI, where ITSA-CFNet is the best method among all. This suggests that RAFT-Stereo already has strong generalization capability.
Combining RAFT-Stereo with NS (group D) consistently produces the best results across the entire Middlebury dataset, and results that are equivalent to RAFT-Stereo trained on synthetic ground-truth on ETH3D, all without requiring any ground-truth data. This results in a small drop in accuracy on KITTI, that is negligible in exchange for the improvement on Middlebury (often 30-40\%). 
%but we consider this negligible in exchange for the significant improvement over self-supervised solutions on Middlebury (often 30-40\%). 

%-------------------------------------------------------------------------

\section{Conclusion}

We have presented a pioneering pipeline that leverages NeRF to train deep stereo networks without the requirement of ground-truth depth or stereo cameras. By capturing images with a single low-cost handheld camera, we generate thousands of stereo pairs for training through our NS paradigm. This approach results in state-of-the-art zero-shot generalization, surpassing both self-supervised and supervised methods. Our work represents a significant advancement towards data democratization, putting the key to the success into the users' hands.

\textbf{Limitations.} Samples collected so far are limited to small-scale, static scenes.
Moreover, our NS -- and any -- stereo networks still fail in some challenging conditions, e.g. transparent surfaces \cite{zamaramirez2022booster} or nighttime images \cite{Gehrig21ral,Argoverse}.
A larger-scale collection campaign, coupled with other NeRFs variants \cite{Mildenhall_2022_CVPR,song2022nerfplayer}, may deal with them in the future.

%\textbf{Ethical Remarks.} The images we collected do not contain any persons nor personal identifiable information.  

\textbf{Future Research.} Our NS pipeline can possibly be extended to generate labels for other dense, low-level tasks such as optical flow (similarly to \cite{Aleotti_CVPR_2021}) or multi-view stereo.

%%%%%%%%% REFERENCES
{\small
\bibliographystyle{ieee_fullname}
\bibliography{reference}
}

\end{document}